\documentclass[10pt,twocolumn,letterpaper]{article}

\usepackage{iccv}
\usepackage{times}
\usepackage{epsfig}
\usepackage{graphicx}
\usepackage{amsmath}
\usepackage{amssymb}
\usepackage{float}                  
\usepackage{multirow,bigdelim}
\usepackage{bm}
\usepackage{nicefrac}
\usepackage{booktabs}
\usepackage{nicefrac}
\usepackage[table]{xcolor}
\usepackage{subcaption}

\usepackage[breaklinks=true,bookmarks=false, hidelinks=True]{hyperref}

\hypersetup{
  pdfinfo={
  pdfproducer={},
  Title={Stochastic Filter Groups for Multi-Task CNNs: Learning Specialist and Generalist Convolution Kernels},
  Subject={},
  Author={Felix J.S. Bragman},
  }
}

\iccvfinalcopy %

\ificcvfinal\fi
\begin{document}

\title{Stochastic Filter Groups for Multi-Task CNNs:\\Learning Specialist and Generalist Convolution Kernels}
\author{Felix J.S. Bragman\thanks{Both authors contributed equally}\\
University College London, UK\\
{\tt\small f.bragman@ucl.ac.uk}
\and
Ryutaro Tanno\footnotemark[1] \\
University College London, UK\\
{\tt\small ryutaro.tanno.15@ucl.ac.uk}
\and
Sebastien Ourselin \\
Kings College London\\
{\tt\small sebastien.ourselin@kcl.ac.uk}
\and
Daniel C. Alexander \\
University College London \\
{\tt\small d.alexander@ucl.ac.uk}
\and
M. Jorge Cardoso \\
Kings College London \\
{\tt\small m.jorge.cardoso@kcl.ac.uk}
\vspace{-0.4in}
}

\vspace{-5mm}
\maketitle

\newcommand{\felix}[1]{\textcolor{black}{#1}}
\newcommand{\ryu}[1]{\textcolor{black}{#1}}

\vspace{-7mm}
\begin{abstract}
\vspace{-3mm}
The performance of multi-task learning in Convolutional Neural Networks (CNNs) hinges on the design of feature sharing between tasks within the architecture. The number of possible sharing patterns are combinatorial in the depth of the network and the number of tasks, and thus hand-crafting an architecture, purely based on the human intuitions of task relationships can be time-consuming and suboptimal. In this paper, we present a probabilistic approach to learning task-specific and shared representations in CNNs for multi-task learning. Specifically, we propose ``stochastic filter groups'' (SFG), a mechanism to assign convolution kernels in each layer to ``specialist'' or ``generalist'' groups, which are specific to or shared across different tasks, respectively. The SFG modules determine the connectivity between layers and the structures of task-specific and shared representations in the network. We employ variational inference to learn the posterior distribution over the possible grouping of kernels and network parameters. Experiments demonstrate that the proposed method generalises across multiple tasks and shows improved performance over baseline methods. 

\end{abstract}
\vspace{-5mm}

\section{Introduction}
\vspace{-2mm}

Multi-task learning (MTL) aims to enhance learning efficiency and predictive performance by simultaneously solving multiple related tasks \cite{caruana1997multitask}. Recently, applications of convolutional neural networks (CNNs) in MTL have demonstrated promising results in a wide-range of computer vision applications, ranging from visual scene understanding \cite{sermanet2014overfeat,eigen2015predicting,MisraCrossMTL16,kokkinos2017ubernet,ranjan2019hyperface,bilen2016integrated} to medical image computing \cite{moeskops2016deep,chen2016bridging,bragman2018multi,tanno2018autodvt}. 

\begin{figure}[t]
\centering 
\vspace{-5mm}
\includegraphics[width=0.4\textwidth]{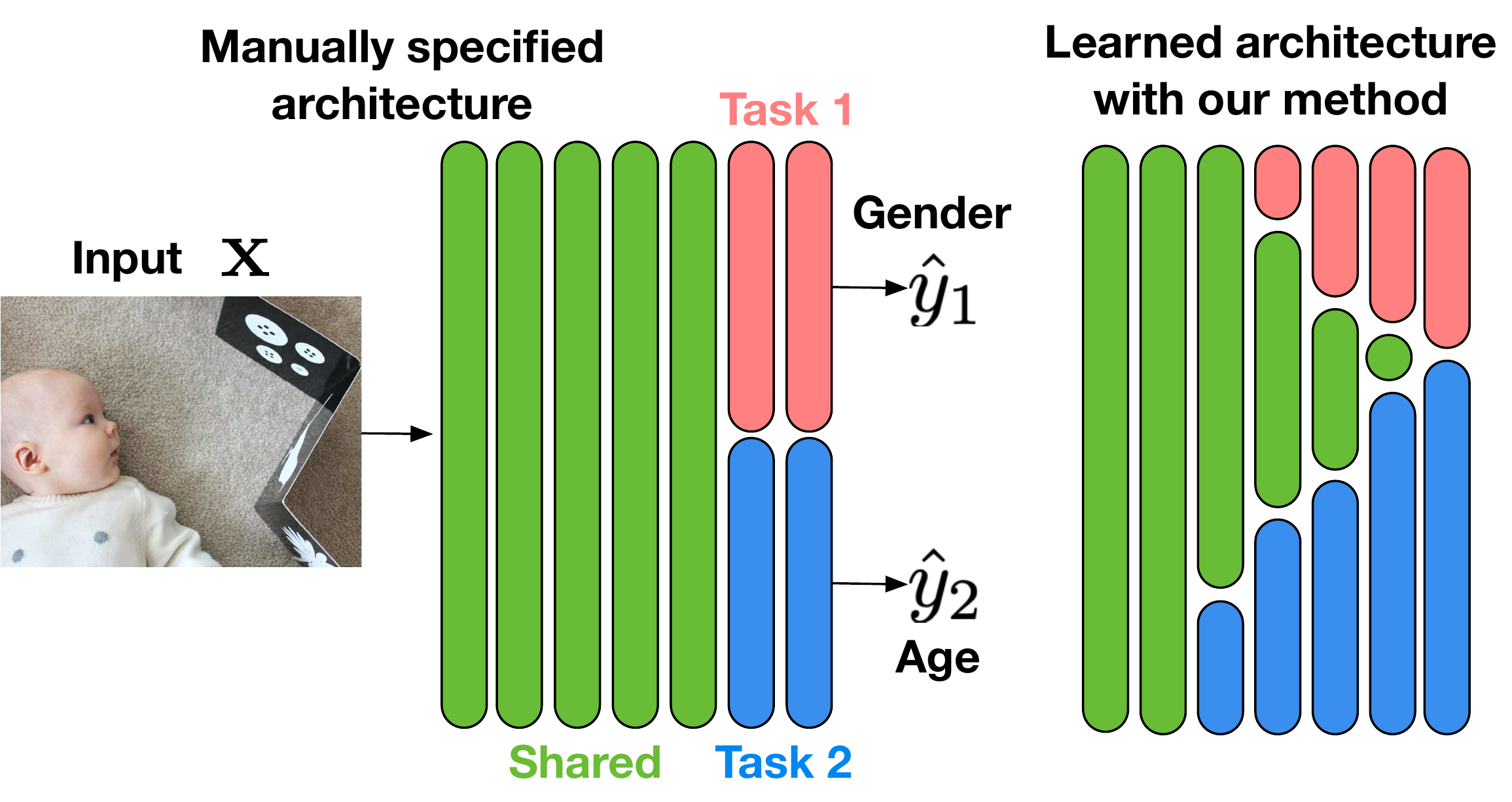}
\vspace{-4mm}
\caption{\small Figure on the left illustrates a typical multi-task architecture, while the figure on the right shows an example architecture that can be learned with our method. We propose \emph{Stochastic Filter Groups}, a principled way to learn the assignment of convolution kernels to task-specific and shared groups. }
\label{fig:intro}
\vspace{-4mm}
\end{figure}

A key factor for successful MTL neural network models is the ability to learn shared and task-specific representations \cite{MisraCrossMTL16}. A mechanism to understand the commonalities and differences between tasks allows the model to transfer information between tasks while tailoring the predictive model to describe the distinct characteristics of the individual tasks. The quality of such representations is determined by the architectural design of where model components such as features \cite{Ruder2019SluiceNL} and weights \cite{meyerson2018beyond} are shared and separated between tasks. However, the space of possible architectures is combinatorially large, and the manual exploration of this space is inefficient and subject to human biases. For example, Fig.~\ref{fig:intro} shows a typical CNN architecture for MTL comprised of a shared ``trunk'' feature extractor and task-specific ``branch'' networks \cite{tanno2018autodvt,huang2015cross,jou2016deep,kendall2017multi,ranjan2019hyperface, bragman2018multi}. The desired amount of shared and task-specific representations, and their interactions within the architecture are dependent on the difficulty of the individual tasks and the relation between them, neither of which are a priori known in most cases \cite{taskonomy2018}. This illustrates the challenge of handcrafting an appropriate architecture, and the need for an effective automatic method to learn it from data.

In this paper, we propose \textit{Stochastic Filter Groups} (SFGs); a probabilistic mechanism to learn the amount of task-specific and shared representations needed in each layer of MTL architectures (Fig.~\ref{fig:intro}). Specifically, the SFGs learns to allocate kernels in each convolution layer into either ``specialist'' groups or a ``shared'' trunk, which are specific to or shared across different tasks, respectively (Fig.~\ref{fig:sfg}). The SFG equips the network with a mechanism to learn inter-layer connectivity and thus the structures of task-specific and shared representations. We cast the learning of SFG modules as a variational inference problem.

We evaluate the efficacy of SFGs on a variety of tasks. In particular, we focus on two multi-task learning problems: 1) age regression and gender classification from face images on UTKFace dataset \cite{zhifei2017cvpr} and 2) semantic regression (i.e. image synthesis) and semantic segmentation on a real-world medical imaging dataset, both of which require predictions over all pixels. Experiments show that our method achieves considerably higher prediction accuracy than baselines with no mechanism to learn connectivity structures, and either higher or comparable performance than a cross-stitch network \cite{MisraCrossMTL16}, while being able to learn meaningful architectures automatically.

\section{Related works}
\vspace{-2mm}
Our work is concerned with the goal of learning where to share neural network components across different tasks to maximise the benefit of MTL. The main challenge of such methods lies in designing a mechanism that determines how and where to share weights within the network. There are broadly two categories of methods that determine the nature of weight sharing and \ryu{separation} in MTL networks.

The first category is composed of methods that optimise the \ryu{structures of weight sharing} in order to maximise task-wise performance. These methods set out to learn a set a vectors that control which features are shared within a layer and how these are distributed across \cite{long2017learning, meyerson2018beyond, MisraCrossMTL16, Ruder2019SluiceNL}. They start with a baseline CNN architecture where they learn additional connections and pathways that define the final MTL model. For instance, Cross-Stitch networks \cite{MisraCrossMTL16} control the degree of weight sharing at each convolution layer whilst Soft-Layer Ordering \cite{meyerson2018beyond} goes beyond the assumption of parallel ordering of feature hierarchies to allow features to mix at different layers depending on the task. \ryu{Routing net \cite{rosenbaum2017routing} proposes an architecture in which each layer is a set of function blocks, and learns to decide which composition of blocks to use given an input and a task.}

The second group of MTL methods focuses on weight clustering based on task-similarity \cite{xue2007multi, jacob2009, Kang2011,lu2017fully,mejjati2018multi}. For example, \cite{lu2017fully} employed an iterative algorithm to grow a tree-like deep architecture that clusters similar tasks hierarchically or \cite{mejjati2018multi} which determines the degree of weight sharing based on statistical dependency between tasks. 

Our method falls into first category, and differentiates itself by performing ``hard' partitioning of task-specific and shared features. By contrast, prior methods are based on ``soft'' sharing of features \cite{MisraCrossMTL16,Ruder2019SluiceNL} or weights \cite{long2017learning,meyerson2018beyond}. These methods generally learn a set of mixing coefficients that determine the weighted sum of features throughout the network, which does not impose connectivity structures on the architecture. On the other hand, our method learns a distribution over the connectivity of layers by grouping kernels. This allows our model to learn meaningful grouping of task-specific and shared features as illustrated in Fig.~\ref{fig:activations}.

\section{Methods}
We introduce a new approach for determining where to learn task-specific and shared representation in multi-task CNN architectures. We propose \textit{stochastic filter groups} (SFG), a probabilistic mechanism to partition kernels in each convolution layer into ``specialist" groups or a ``shared" group, which are specific to or shared across different tasks, respectively. We employ variational inference to learn the distributions over the possible grouping of kernels and network parameters that determines the connectivity between layers and the shared and task-specific features. This naturally results in a learning algorithm that optimally allocate representation capacity across multi-tasks via gradient-based stochastic optimization, e.g. stochastic gradient descent.

\begin{figure}[ht]
	\center
	\includegraphics[width=0.95\linewidth]{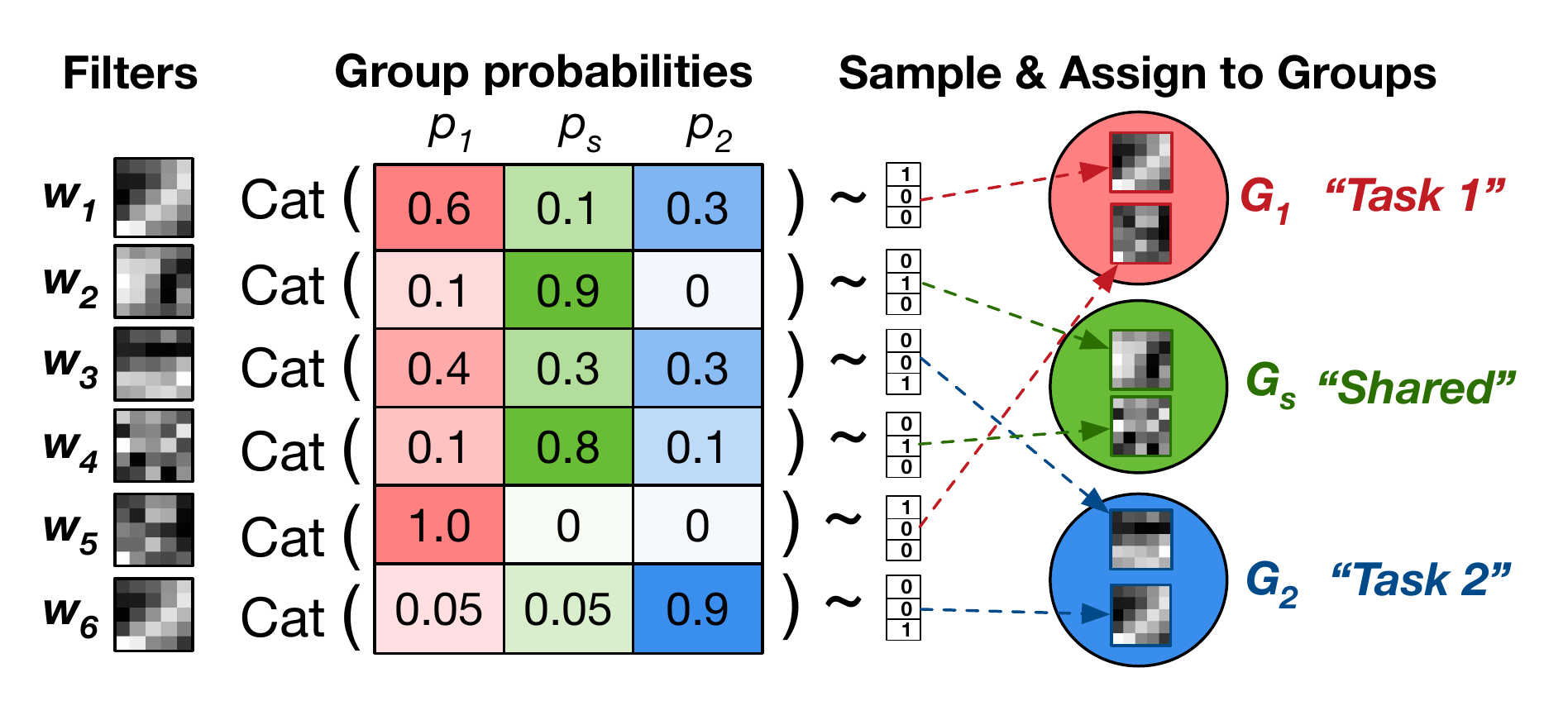}
	\caption{\small Illustration of filter assignment in a SFG module. Each kernel $\{\mathbf{w}_{k}\}$ in the given convolution layer is probabilistically assigned to one of the filter groups $G_1, G_{s}, G_{2}$ according to the sample drawn from the associated categorical distribution $\text{Cat}(p_1, p_{s}, p_2)$. }
    \label{fig:sfg}
\end{figure}

\begin{figure}[ht]
	\center
 	\vspace{-5mm}
	\includegraphics[width=\linewidth]{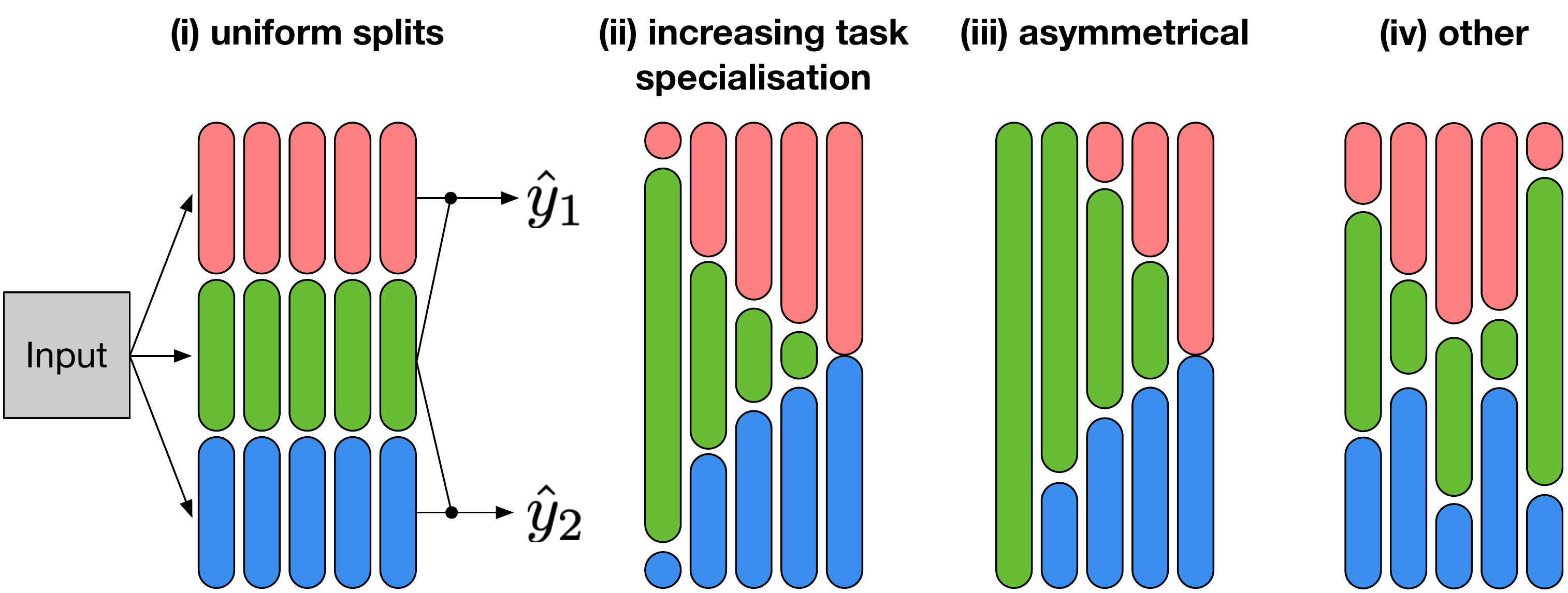}
	\caption{\small Illustration of possible grouping patterns learnable with the proposed method. Each set of green, pink and yellow blocks represent the ratio of filter groups $G_1$ (red), $G_{s}$ (green) and $G_{2}$ (blue). (i) denotes the case where all kernels are uniformly split. (ii) \& (iii) are the cases where the convolution kernels become more task-specific at deeper layers. (iv) shows an example with more heterogeneous splits across tasks.}
    \label{fig:different_grouping}
\end{figure}

\begin{figure}[ht]
	\center
	\includegraphics[width=0.9\linewidth]{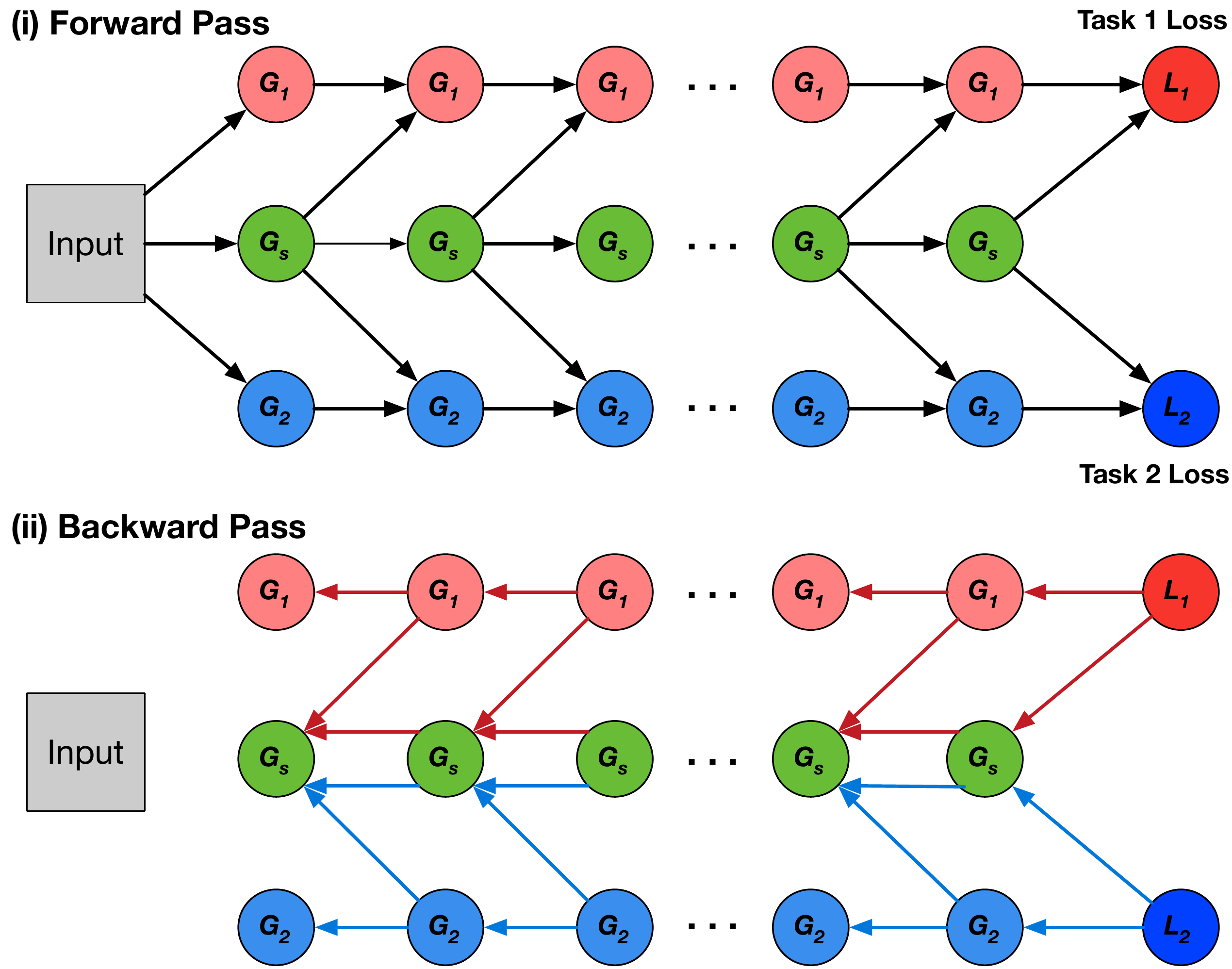}
	\caption{\small Illustration of feature routing. The circles $G_{1}, G_{s}, G_{2}$ denote the task-specific and shared filter groups in each layer. (i) shows the directions of routing of activations between different filter groups while (ii) shows the directions of the gradient flow from the task losses $L_{1}$ and $L_{2}$. The red and blue arrows denote the gradients that step from $L_{1}$ and $L_{2}$, respectively. The task-specific groups $G_{1}, G_{2}$ are only updated based on the associated losses, while the shared group $G_{s}$ is updated based on both. }
    \label{fig:forward_and_backward}
\end{figure}

\begin{figure*}[ht]

    \center
	\includegraphics[width=0.95\linewidth]{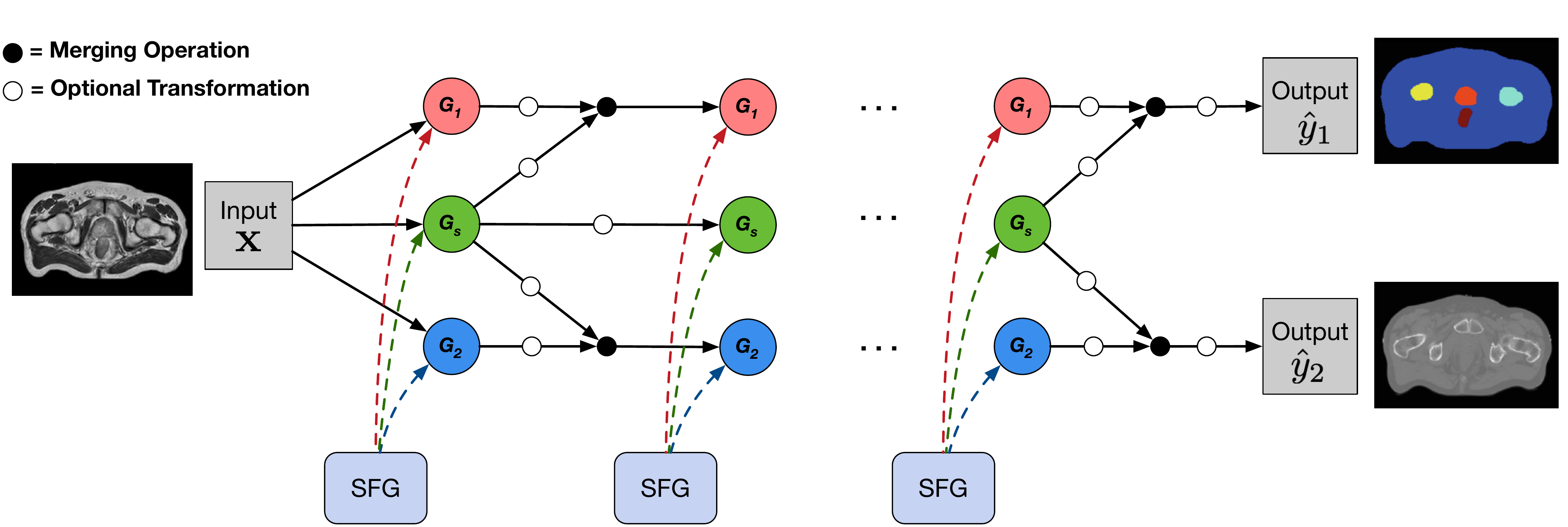}
	\caption{\small Schematic of the proposed multi-task architecture based on a series of SFG modules in the presence of two tasks. At each convolution layer, kernels are stochastically assigned to task-specific and shared filter groups $G_{1}, G_{s}, G_{2}$. Each input image is first convolved with the respective filter groups to yield three distinct sets of output activations, which are routed sparsely to the filter groups in the second layer layer. This process repeats in the remaining SFG modules in the architecture until the last layer where the outputs of the final SFG module are combined into task-specific predictions $\hat{y}_{1}$ and $\hat{y}_{2}$. Each small white circle denotes an optional transformation (e.g. extra convolutions) and black circle merges the incoming inputs (e.g. concatenation).}
    \label{fig:schematic}
\end{figure*}

\subsection{Stochastic Filter Groups}
SFGs introduce a sparse connection structure into the architecture of CNN for multi-task learning in order to separate features into task-specific and shared components. Ioannou et al. \cite{ioannou2017deep} introduced \textit{filter groups} to partition kernels in each convolution layer into groups, each of which acts only on a subset of the preceding features. They demonstrated that such sparsity reduces computational cost and number of parameters without compromising accuracy. Huang et al. \cite{huang2018condensenet} proposed a similar concept, but differs in that the \textit{filter groups} do not operate on mutually exclusive sets of features. Here we adapt the concept of filter groups to the multi-task learning paradigm and propose an extension with an additional mechanism for learning an optimal kernel grouping rather than pre-specifying them.

For simplicity, we describe SFGs for the case of multitask learning with two tasks, but can be trivially extended to a larger number of tasks. At the $l^{\text{th}}$ convolution layer in a CNN architecture with $K_l$ kernels $\{\mathbf{w}^{(l),k}\}^{K_l}_{k=1}$, the associated SFG performs two operations: 

\begin{enumerate}
    \item \textbf{Filter Assignment:} each kernel $\mathbf{w}_{k}^{(l)}$ is stochastically assigned to either: i) the ``task-1 specific group'' $G^{(l)}_{1}$, ii) ``shared group'' $G^{(l)}_{s}$ or iii) ``task-2 specific group'' $G^{(l)}_{2}$ with respective probabilities $\mathbf{p}^{(l),k} = [p^{(l),k}_1, p^{(l),k}_{s}, p^{(l),k}_2] \in [0,1]^3$. Convolving with the respecitve filter groups yields distinct sets of features $F^{(l)}_1, F^{(l)}_{s}, F^{(l)}_{2}$. Fig.~\ref{fig:sfg} illustrates this operation and Fig.~\ref{fig:different_grouping} shows different learnable patterns. 
    
    \item \textbf{Feature Routing:} as shown in Fig.~\ref{fig:forward_and_backward}~(i), the features $F^{(l)}_1, F^{(l)}_{s}, F^{(l)}_{2}$ are routed to the filter groups $G_{1}^{(l+1)}, G_{s}^{(l+1)}, G_{2}^{(l+1)}$ in the subsequent $(l+1)^{\text{th}}$ layer in such a way to respect the task-specificity and sharedness of filter groups in the $l^{\text{th}}$ layer. Specifically, we perform the following routing for $l>0$: 
        \begin{align*}
        F^{(l+1)}_1 &= h^{(l+1)}\big{(}[F^{(l)}_1|F^{(l)}_{s}]*G^{(l+1)}_{1}\big{)}\\ 
        F^{(l+1)}_{s} &= h^{(l+1)}\big{(} F^{(l)}_{s}*G^{(l+1)}_{s}\big{)}\\ 
        F^{(l+1)}_2 &= h^{(l+1)}\big{(}[F^{(l)}_2|F^{(l)}_{s}]*G^{(l+1)}_{2}\big{)}
        \end{align*}
    where each $h^{(l+1)}$ defines the choice of non-linear function, $*$ denotes convolution operation and $|$ denotes a merging operation of arrays (e.g. concatenation). At $l=0$, input image $\mathbf{x}$ is simply convolved with the first set of filter groups to yield $F^{(1)}_{i} = h^{(1)}\big{(}\mathbf{x}*G^{(1)}_{i}\big{)}, i\in\{1,2,s\}$. Fig.~\ref{fig:forward_and_backward}(ii) shows that such sparse connectivity ensures the parameters of $G^{(l)}_{1}$ and $G^{(l)}_{2}$ are only learned based on the respective task losses, while $G^{(l)}_{s}$ is optimised based on both tasks. 
\end{enumerate}

Fig.~\ref{fig:schematic} provides a schematic of our overall architecture, in which each SFG module stochastically generates filter groups in each convolution layer and the resultant features are sparsely routed as described above. The merging modules, denoted as black circles, combine the task-specific and shared features appropriately, i.e. $[F^{(l)}_{i}|F^{(l)}_{s}], i = 1,2$ and pass them to the filter groups in the next layer. Each white circle denotes the presence of additional transformations  (e.g. convolutions or fully connected layers) in each $h^{(l+1)}$, performed on top of the standard non-linearity (e.g. ReLU).

The proposed sparse connectivity is integral to ensure task performance and structured representations. In particular, one might argue that the routing of ``shared'' features $F^{(l)}_{s}$ to the respective ``task-specific'' filter groups $G^{(l+1)}_{1}$ and $G^{(l+1)}_{2}$ is not necessary to ensure the separation of gradients across the task losses. However, this connection allows for learning more complex task-specific features at deeper layers in the network. For example, without this routing, having a large proportion of ``shared'' filter group $G_{s}$ at the first layer (Fig.~\ref{fig:different_grouping}~(ii)) substantially reduces the amount of features available for learning task-specific kernels in the subsequent layers---in the extreme case in which all kernels in one layer are assigned to $G_{s}$, the task-specific filter groups in the subsequent layers are effectively unused. 

Another important aspect that needs to be highlighted is the varying dimensionality of feature maps. Specifically, the number of kernels in the respective filter groups $G^{(l)}_{1}, G^{(l)}_{s}, G^{(l)}_{2}$ can vary at each iteration of the training, and thus, so does the depth of the resultant feature maps $F^{(l)}_{1}, F^{(l)}_{s}, F^{(l)}_{2}$. Instead of directly working with features maps of varying size, we implement the proposed architecture by defining $F^{(l)}_{1}, F^{(l)}_{s}, F^{(l)}_{2}$ as sparse tensors. At each SFG module, we first convolve the input features with all kernels, and generate the output features from each filter group by zeroing out the channels that root from the kernels in the other groups, resulting in $F^{(l)}_{1}, F^{(l)}_{s}, F^{(l)}_{2}$ that are sparse at non-overlapping channel indices. In the simplest form with no additional transformation (i.e. the grey circles in Fig.~\ref{fig:schematic} are identity functions), we define the merging operation $[F^{(l)}_{i}|F^{(l)}_{s}], i = 1,2$ as pixel-wise summation. In the presence of more complex transforms (e.g. residual blocks), we concatenate the output features in the channel-axis and perform a 1x1 convolution to ensure the number of channels in $[F^{(l)}_{i}|F^{(l)}_{s}]$ is the same as in $F^{(l)}_{s}$.

\subsection{T+1 Way Concrete ``Drop-Out''}
Here we derive the method for simultaneously optimising the CNN parameters and grouping probabilities. We achieve this by extending the variational interpretation of binary dropout \cite{gal2016uncertainty,gal2017concrete} to the $(T+1)$-way assignment of each convolution kernel to the filter groups where $T$ is the number of tasks. As before, we consider the case $T=2$. 

Suppose that the architecture consists of $L$ SFG modules, each with $K_l$ kernels where $l$ is the index. As the posterior distribution over the convolution kernels in SFG modules $p(\mathcal{W}|\textbf{X}, \mathbf{Y}^{(1)}, \mathbf{Y}^{(2)})$ is intractable, we approximate it with a simpler distribution $q_{\phi}(\mathcal{W})$ where $\mathcal{W}=\{\mathbf{W}^{(l),k}\}_{k=1,...,K_{l},l=1,...,L}$. Assuming that the posterior distribution factorizes over layers and kernels up to group assignment, we defined the variational distribution as:
\begin{align*}
 q_{\phi}(\mathcal{W}) &= \prod_{l=1}^{L}\prod_{k=1}^{K_{l}} q_{\phi_{lk}}(\mathbf{W}^{(l),k}) \\
 &= \prod_{l=1}^{L}\prod_{k=1}^{K_{l}} q_{\phi_{lk}}(\mathbf{W}^{(l),k}_1,\mathbf{W}^{(l),k}_{s}, \mathbf{W}^{(l),k}_{2})
\end{align*}
where $\{\mathbf{W}^{(l),k}_1,\mathbf{W}^{(l),k}_{s}, \mathbf{W}^{(l),k}_{2}\}$ denotes the $k^{\text{th}}$ kernel in $l^{\text{th}}$ convolution layer after being routed into task-specific $G^{(l)}_1, G^{(l)}_2$ and shared group $G^{(l)}_{s}$. We define each $q_{\phi_{lk}}(\mathbf{W}^{(l),k}_1,\mathbf{W}^{(l),k}_2, \mathbf{W}^{(l),k}_{s})$ as: 
\begin{align}
\mathbf{W}^{(l),k}_{i} &= z^{(l),k}_{i}\cdot\mathbf{M}^{(l),k}\, \,\,\text{for } i \in\{1,s,2\}\\
\mathbf{z}^{(l),k}&=[z^{(l),k}_{1}, z^{(l),k}_{2},  z^{(l),k}_{s}] \sim \text{Cat}(\mathbf{p}^{(l),k}) \label{eq:sample_cat}
\end{align}
where $\mathbf{z}^{(l),k}$ is the one-hot encoding of a sample from the categorical distribution over filter group assignments, and $\mathbf{M}^{(l),k}$ denotes the parameters of the pre-grouping convolution kernel. The set of variational parameters for each kernel in each layer is thus given by $\phi_{lk} = \{\mathbf{M}^{(l),k}, \mathbf{p}^{(l),k}=[p^{(l),k}_1, p^{(l),k}_{s}, p^{(l),k}_2]\}$. 

We minimize the KL divergence between the approximate posterior $q_{\phi}(\mathcal{W})$ and $p(\mathcal{W}|\textbf{X}, \mathbf{Y}^{(1)}, \mathbf{Y}^{(2)})$. Assuming that the joint likelihood over the two tasks factorizes, we have the following optimization objective:
\begin{multline}\label{eq:variational_loss}
    \mathcal{L}_{\text{MC}}(\phi) = -\frac{N}{M}\sum_{i=1}^{M} \Big{[}\text{log } p(y^{(1)}_i|\mathbf{x}_i, \mathcal{W}_i) + \text{log }p(y^{(2)}_i|\mathbf{x}_i, \mathcal{W}_i)\Big{]} \\ + \sum_{l=1}^{L}\sum_{k=1}^{K_l}\text{KL}(q_{\phi_{lk}}(\mathbf{W}^{(l),k})||p(\mathbf{W}^{(l),k}))
\end{multline}
where $M$ is the size of the mini-batch, $N$ is the total number of training data points, and $\mathcal{W}_i$ denotes a set of model parameters sampled from $q_{\phi}(\mathcal{W})$. The last KL term regularizes the deviation of the approximate posterior from the prior $p(\mathbf{W}^{(l),k})=\mathcal{N}(0, \mathbf{I}/l^{2})$ where $l>0$. Adapting the approximation presented in \cite{gal2016uncertainty} to our scenario, we obtain:
\begin{equation}\label{eq:KL}
    \text{KL}(q_{\phi_{lk}}(\mathbf{W}^{(l),k})||p(\mathbf{W}^{(l),k})) \propto \frac{l^2}{2}||\mathbf{M}^{(l),k}||^{2}
    _{2} - \mathcal{H}(\mathbf{p}^{(l),k})
\end{equation}
where $\mathcal{H}(\mathbf{p}^{(l),k})=-\sum_{i\in \{1,2,s\}}p^{(l),k}_i\text{log }p^{(l),k}_i$ is the entropy of the grouping probabilities. While the first term performs the L2-weight norm, the second term pulls the grouping probabilities towards the uniform distribution. Plugging eq.\eqref{eq:KL} into eq.\eqref{eq:variational_loss} yields the overall loss: 

\begin{multline}\label{eq:total_loss}
\medmuskip=0mu
\thinmuskip=0mu
\thickmuskip=0mu
\mathcal{L}_{\text{MC}}(\phi) = -\frac{N}{M}\sum_{i=1}^{M} \left[\log\text{\,} p\left(y^{(1)}_i|\mathbf{x}_i, \mathcal{W}_i\right) + \log\text{\,} p\left(y^{(2)}_i|\mathbf{x}_i, \mathcal{W}_i\right)\right] \\ + \lambda_{1} \cdot \sum_{l=1}^{L}\sum_{k=1}^{K_l}||\mathbf{M}^{(l),k}||^{2} - \lambda_{2} \cdot \sum_{l=1}^{L}\sum_{k=1}^{K_l}\mathcal{H}(\mathbf{p}^{(l),k})
\end{multline}
where $\lambda_1>0, \lambda_2>0$ are regularization coefficients. 

We note that the discrete sampling operation during filter group assignment (eq.~\eqref{eq:sample_cat}) creates discontinuities, giving the first term in the objective function (eq.~\ref{eq:total_loss}) zero gradient with respect to the grouping probabilities $\{\mathbf{p}^{(l),k}\}$. We therefore, as employed in \cite{kendall2017multi} for the binary case, approximate each of the categorical variables $\text{Cat}(\mathbf{p}^{(l),k})$ by the Gumbel-Softmax distribution, $\text{GSM}(\mathbf{p}^{(l),k}, \tau)$ \cite{maddison2016concrete,jang2016categorical}, a continuous relaxation which allows for sampling, differentiable with respect to the parameters $\mathbf{p}^{(l),k}$ through a reparametrisation trick. The temperature term $\tau$ adjusts the bias-variance tradeoff of gradient approximation; as the value of $\tau$ approaches 0, samples from the GSM distribution become one-hot (i.e. lower bias) while the variance of the gradients increases. \textcolor{black}{In practice, we start at a high $\tau$ and anneal to a small but non-zero value as in \cite{jang2016categorical,gal2017concrete} as detailed in supplementary materials.}

\section{Experiments}\label{sec:experiments}
We tested \emph{stochastic filter groups} (SFG) on two multi-task learning (MTL) problems: 1) age regression and gender classification from face images on UTKFace dataset \cite{zhifei2017cvpr} and 2) semantic image regression (synthesis) and segmentation on a medical imaging dataset. Full details of the training and datasets are provided in Sec.~A in the supplementary materials. 
    \paragraph{UTKFace dataset:} We tested our method on UTKFace \cite{zhifei2017cvpr}, which consists of 23,703 cropped faced images in the wild with labels for age and gender. We created a dataset with a 70/15/15\% split. We created a secondary separate dataset containing only 10\% of images from the initial set, so as to simulate a data-starved scenario.
    \paragraph{Medical imaging dataset:} We used a medical imaging dataset to evaluate our method in a real-world, multi-task problem where paucity of data is common and hard to mitigate. The goal of radiotherapy treatment planning is to maximise radiation dose to the tumour whilst minimising dose to the organs. To plan dose delivery, a Computed Tomography (CT) scan is needed as CT voxel intensity scales with tissue density, thus allowing dose propagation simulations. An MRI scan is needed to segment the surrounding organs. Instead of acquiring both an MRI and a CT, algorithms can be used to synthesise a CT scan (task 1) and segment organs (task 2) given a single input MRI scan. For this experiment, we acquired $15$\felix{,} 3D prostate cancer scans with respective CT and MRI scans with semantic 3D labels for organs (prostate, bladder, rectum and left/right femur heads) obtained from a trained radiologist. We created a training set of $10$ patients, with the remaining $5$ used for testing. We trained our networks on 2D \felix{patches} of size $128$x$128$ randomly sampled from axial slices, and reconstructed the 3D volumes of size $288$x$288$x$62$ at test time by stitching together the subimage-wise predictions.

\subsection{Baselines}
We compared our model against four baselines in addition to Cross-Stitch networks \cite{MisraCrossMTL16} trained end-to-end rather than sequentially for fair comparison. The four baselines considered are: 1) single-task networks, 2) hard-parameter sharing multi-task network (MT-hard sharing), 3) SFG-networks with constant $\nicefrac{1}{3}$ allocated grouping (MT-constant mask) as \textit{per} Fig.~\ref{fig:different_grouping}(i), and 4) SFG-networks with constant grouping probabilities (MT-constant \textbf{p}). We train all the baselines in an end-to-end fashion for all the experiments. 

We note that all four baselines can be considered special cases of an SFG-network. Two \emph{single-task networks} can be learned when the shared grouping probability of kernels is set to zero. Considering Fig.~\ref{fig:schematic}, this would remove the diagonal connections and the shared network. This may be important when faced with two unrelated tasks which share no contextual information. A \emph{hard-parameter sharing network} exists when all shared grouping probabilities are maximised to one leading to a scenario where all features are shared within the network up until the task-specific layers. The \emph{MT-constant mask network} is illustrated in Fig.~ \ref{fig:different_grouping}(i), where $\nicefrac{1}{3}$ of kernels are allocated to the task $1$, task $2$ and shared groups, yielding uniform splits across layers. This occurs when an equal number of kernels in each layer obtain probabilities of $\mathbf{p}^{(l),k}=[1, 0, 0], [0, 1, 0]$ and $[0, 0, 1]$. Lastly, the \emph{MT-constant \textbf{p}} model represents the situation where the grouping is non-informative and each kernel has equal probability of being specific or shared with probability $\mathbf{p}^{(l),k}=[\nicefrac{1}{3}, \nicefrac{1}{3}, \nicefrac{1}{3}]$. \textcolor{black}{Training details for these models, including the hyper-parameter settings, are provided in Sec.~B in the supplementary document.}

    \paragraph{UTKFace network:} We used VGG-11 CNN architecture \cite{vgg} for age and gender prediction. The network consists of a series of $3$x$3$ convolutional layers interleaved with max pooling layers. In contrast to the original architecture, we replaced the final max pooling and fully connected layers with global average pooling (GAP) followed by a fully connected layers for prediction. Our model's version of VGG (SFG-VGG) replaces each convolutional layer in VGG-11 with a SFG layer with max pooling applied to each feature map $F^{(l)}_{1}$, $F^{(l)}_{2}$, $F^{(l)}_{s}$. We applied GAP to each final feature map before the final merging operation and two fully connected layers for each task.
    
    \paragraph{Medical imaging network:} We used the HighResNet architecture \cite{wenqi} for CT synthesis and organ segmentation. This network has been developed for semantic segmentation in medical imaging and has been used in a variety of medical applications such as CT synthesis \cite{bragman2018multi} and brain segmentation \cite{wenqi}. It consists of a series of residual blocks, which group two $3$x$3$ convolutional layers with dilated convolutions. The baseline network is composed of a $3$x$3$ convolutional layer followed by three sets of twice repeated residual blocks with dilated convolutions using factors $d=[1, 2, 4]$. There is a $3$x$3$ convolutional layer between each set of repeated residual blocks. The network ends with two final $3$x$3$ layers and either one or two $1$x$1$ convolutional layers for single and multi-task predictions. In our model, we replace each convolutional layer with an SFG module. After the first SFG layer, three distinct repeated residual blocks are applied to $F^{(l=0)}_{1}$, $F^{(l=0)}_{2}$, $F^{(l=0)}_{s}$. These are then merged according the feature routing methodology followed by a new SFG-layer and subsequent residual layers. Our model concludes with 2 successive SFG-layers followed by $1$x$1$ convolutional layers applied to the merged features $F^{(l=L)}_{1}$ and $F^{(l=L)}_{2}$.\looseness=-1

\begin{table}[h]
    \begin{subtable}[t]{0.9\linewidth}
    \small
    \caption{Full training data}
    \vspace{-5mm}
    \begin{center}
        \begin{tabular}{lccc}
        \hline
            \toprule
            \multirow{2}{*}{Method}                      & Age   & Gender     \\
                                                        & (MAE)   & (Accuracy)    \\
            \midrule
            One-task (VGG11) \cite{vgg}                       & $7.32$  & $90.70$                 \\
            \multirow{1}{*}{MT-hard sharing}   & $7.92$  & $90.60$ \\ 
            \multirow{1}{*}{MT-constant mask}   & $7.67$  & $89.41$                 \\
            \multirow{1}{*}{MT-constant \textbf{p}=[$\nicefrac{1}{3}$,$\nicefrac{1}{3}$,$\nicefrac{1}{3}$]}   & \cellcolor{blue!15} $6.34$  & \cellcolor{blue!15} $92.10$                 \\
            \multirow{1}{*}{VGG11 Cross Stitch \cite{MisraCrossMTL16}}   & $6.78$  & $90.30$                 \\
            \multirow{1}{*}{MT-SFG (ours)}       &  \cellcolor{red!15} $\mathbf{6.00}$  &  \cellcolor{red!15} $\mathbf{92.46}$                 \\
        \hline
        \end{tabular}
    \end{center}
    \label{tab:big_data}
    \end{subtable}
    \begin{subtable}[t]{0.9\linewidth}
    \small
    \vspace{2mm}
    \caption{Small training data}
    \vspace{-5mm}
    \begin{center}
        \begin{tabular}{lccc}
        \hline
            \toprule
            \multirow{2}{*}{Method} & Age   & Gender     \\
            & (MAE)   & (Accuracy)    \\
            \midrule
            One-task (VGG11) \cite{vgg}                       & \cellcolor{blue!15}  $8.79$  & $85.54$                 \\
            \multirow{1}{*}{MT-hard sharing}   & $9.19$  & $85.83$                 \\
            \multirow{1}{*}{MT-constant mask}   & $9.02$  & $85.98$                \\
            \multirow{1}{*}{MT-constant \textbf{p}=[$\nicefrac{1}{3}$,$\nicefrac{1}{3}$,$\nicefrac{1}{3}$]}   & $9.15$  & \cellcolor{blue!15} $86.01$                 \\
            \multirow{1}{*}{VGG11 Cross Stitch \cite{MisraCrossMTL16}}   & $8.85$  &   $83.72$                 \\
            
            \multirow{1}{*}{MT-SFG (ours)}       & \cellcolor{red!15}  $\mathbf{8.54}$  &\cellcolor{red!15}  $\mathbf{87.01}$                 \\
    \hline
    \end{tabular}
    \end{center}
    \label{tab:small_data}
    \end{subtable}
\vspace{-3mm}
\caption{\small Age regression and gender classification results on UTKFace \cite{zhifei2017cvpr} with (a) the full and (b) limited training set. The best and the second best results are shown in red and blue. The mean absolute error (MAE) is reported for the age prediction and classification accuracy for gender prediction. For our model, we performed $50$ stochastic forward passes at test time by sampling the kernels from the approximate posterior $q_{\phi}(\mathcal{W})$. We calculated the average age per subject and obtained gender prediction using the mode of the test-time predictions.} %

\label{tab:face_set}
\end{table}

\section{Results}

\subsection{Age regression and gender prediction}

Results on age prediction and gender classification on both datasets are presented in Tab.~ \ref{tab:big_data} and \ref{tab:small_data}. Our model (MT-SFG) achieved the best performance in comparison to the baselines in both data regimes. In both sets of experiments, our model outperformed the hard-parameter sharing (\emph{MT-hard sharing}) and constant allocation (\emph{MT-constant mask}). This demonstrates the advantage of learning to allocate kernels. In the \emph{MT-constant mask} model, kernels are equally allocated across groups. In contrast, our model is able to allocate kernels in varying proportions across different layers in the network (Fig.~ \ref{fig:LEARNED_A} - SFG-VGG11) to maximise inductive transfer. Moreover, our methods performed better than a model with constant, non-informative grouping probabilities (\emph{MT-constant} \textbf{p}$= [\nicefrac{1}{3}, \nicefrac{1}{3}, \nicefrac{1}{3}]$), displaying the importance of learning structured representations and connectivity across layers to yield good predictions.

\begin{table*}[t!]
    \begin{subtable}[t]{1.0\linewidth}
        \caption{CT Synthesis (PSNR)}
        \vspace{-5mm}
        \begin{center}
            \footnotesize
            \begin{tabular}{l|c|ccccc}
            \hline
                \toprule
                Method   & Overall   &  Bones  & Organs &  Prostate & Bladder & Rectum  \\
                \midrule
                 One-task (HighResNet) \cite{wenqi} & $\,\,$ 25.76 (0.80)$\,\,$ & $\,\,$ 30.35 (0.58) $\,\,$  & $\,\,$ 38.04 (0.94) $\,\,$ & $\,\,$ 51.38 (0.79) $\,\,$ & $\,\,$ 33.34 (0.83)$\,\,$ & $\,\,$ 34.19 (0.31)$\,\,$  \\
                \multirow{1}{*}{MT-hard sharing }  & 26.31 (0.76) & 31.25 (0.61) & 39.19 (0.98) & 52.93 (0.95)  & 34.12 (0.82) & 34.15 (0.30) \\
                \multirow{1}{*}{MT-constant mask}   &  $24.43 (0.57)$  & $29.10 (0.46)$  & $37.24 (0.86)$ & $50.48 (0.73) $  & $32.29 (1.01)$ & $33.44 (2.88)$  \\
                \multirow{1}{*}{MT-constant \textbf{p}=[$\nicefrac{1}{3}$,$\nicefrac{1}{3}$,$\nicefrac{1}{3}$]}   & 26.64(0.54)  & 31.05 (0.55) & 39.11 (1.00)    & \cellcolor{blue!15} 53.20 (0.86)  & 34.34 (1.35) &  35.61 (0.35) \\
                \multirow{1}{*}{Cross Stitch \cite{MisraCrossMTL16}}   & \cellcolor{red!15}  27.86 (1.05)  & \cellcolor{blue!15}  32.27 (0.55)  & \cellcolor{red!15}  40.45 (1.27)  & \cellcolor{red!15} 54.51 (1.01)     & \cellcolor{red!15}  36.81 (0.92) &  \cellcolor{red!15} 36.35 (0.38)     \\
                \multirow{1}{*}{MT-SFG (ours)}  &   \cellcolor{blue!15} 27.74 (0.96) &  \cellcolor{red!15}  32.29 (0.59)&  \cellcolor{blue!15}  39.93 (1.09)& 53.01 (1.06)   &  \cellcolor{blue!15} 35.65 (0.44)& \cellcolor{blue!15} 35.65 (0.37)  \\
        \hline
        \end{tabular}
        \end{center}
    \label{tab:ct}
    \end{subtable}
    
    \begin{subtable}[t]{1.0\linewidth}
        \vspace{0mm}
        \caption{Segmentation (DICE)}
        \vspace{-5mm}
        \begin{center}
            \centering
            \footnotesize
            \begin{tabular}{l|c|ccccc}
            \hline
                \toprule
                Method & Overall  &   Left Femur Head &  Right Femur Head &  Prostate & Bladder & Rectum   \\
                
                \midrule
                One-task (HighResNet) \cite{wenqi} & $0.848 (0.024)$ & 0.931 (0.012)  & \cellcolor{blue!15} 0.917 (0.013)   & 0.913 (0.013) & 0.739 (0.060)  & 0.741 (0.011)  \\
                
                \multirow{1}{*}{MT-hard sharing }  & $0.829 (0.023)$ & \cellcolor{blue!15} 0.933 (0.009)  & 0.889 (0.044)    &  0.904 (0.016) & 0.685 (0.036)   &  0.732 (0.014) \\
               
                \multirow{1}{*}{MT-constant mask}  & $0.774 (0.065)$ & 0.908 (0.012)  & 0.911 (0.015)    &0.806 (0.0541)   & 0.583 (0.178)  & 0.662 (0.019) \\
                \multirow{1}{*}{MT-constant \textbf{p}=[$\nicefrac{1}{3}$,$\nicefrac{1}{3}$,$\nicefrac{1}{3}$]}   & $0.752 (0.056)$ &   0.917 (0.004)& \cellcolor{red!15} 0.917 (0.01)&0.729 (0.086)   & 0.560 (0.180)  &0.639 (0.012)\\
                 \multirow{1}{*}{Cross Stitch \cite{MisraCrossMTL16}}  &\cellcolor{red!15}  0.854 (0.036) & 0.923 (0.008)   &  0.915 (0.013) &   \cellcolor{red!15} 0.933 (0.009)  &  \cellcolor{red!15} 0.761 (0.053)   &   \cellcolor{blue!15} 0.737 (0.015)\\
                 \multirow{1}{*}{MT-SFG (ours)}  &  \cellcolor{blue!15}  $0.852 (0.047)$ &  \cellcolor{red!15} 0.935 (0.007)  & 0.912 (0.013)   &  \cellcolor{blue!15} 0.923 (0.016)  &  \cellcolor{blue!15}  0.750 (0.062) &  \cellcolor{red!15} 0.758 (0.011) \\
        \hline
        \end{tabular}
        \end{center}
    \label{tab:seg}
    \end{subtable}

\vspace{-5mm}
\caption{\small Performance on the medical imaging dataset with best results in red, and the second best results in blue. The PSNR is reported for the CT-synthesis (synCT) across the whole volume (overall), at the bone regions, across all organ labels and individually at the prostate, bladder and rectum. For the segmentation, the average DICE score per patient across all semantic labels is computed. The standard deviations are computed over the test subject cohort. For our model, we perform $50$ stochastic forward passes at test-time by sampling the kernels from the approximated posterior distribution $q_{\phi}(\mathcal{W})$. We compute the average of all passes to obtain the synCT and calculate the mode of the segmentation labels for the final segmentation.} %
\label{tab:medicalimaging}
\vspace{-2mm}
\end{table*}

\subsection{Image regression and semantic segmentation}
Results on CT image synthesis and organ segmentation from input MRI scans is detailed in Tab.~\ref{tab:medicalimaging}. Our method obtains equivalent (non-statistically significant different) results to the Cross-Stitch network \cite{MisraCrossMTL16} on both tasks. We have, however, observed best synthesis performance in the bone regions (femur heads and pelvic bone region) in our model when compared against all the baselines, including Cross-Stitch. The bone voxel intensities are the most difficult to synthesise from an input MR scan as task uncertainty in the MR to CT mapping at the bone is often highest \cite{bragman2018multi}. Our model was able to disentangle features specific to the bone intensity mapping (Fig.~ \ref{fig:activations}) without supervision of the pelvic location, which allowed it to learn a more accurate mapping of an intrinsically difficult task.

\vspace{-2mm} 
\subsection{Learned architectures}
\begin{figure}[b!]
	\center
	\vspace{-4mm}
	\includegraphics[width=1.0\linewidth]{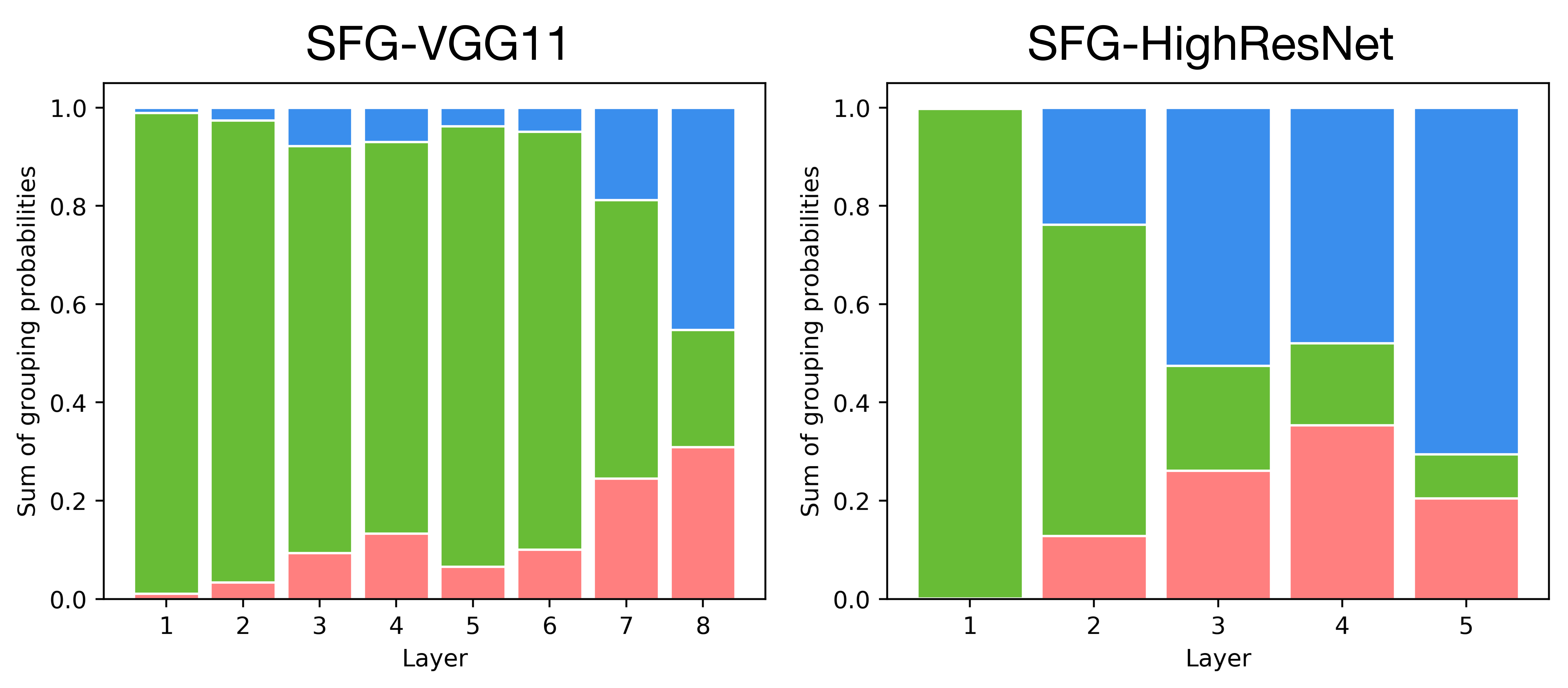}
	\vspace{-7mm}
	\caption{\small Learned kernel grouping in a) SFG-VGG11 network on UTKFace and b) SFG-HighResNet on medical scans. The proportions of task-1, shared and task-2 filter groups are shown in blue, green and pink. Within SFG-VGG11, task-1 age regression and task-2 is gender classification. For SFG-HighResNet, task-1 is CT synthesis and task-2 is organ segmentation. }
	\vspace{-3mm}
    \label{fig:LEARNED_A}
\end{figure}

\begin{figure*}[ht]
 	\vspace{-5mm}
    \center
	\includegraphics[width=0.90\linewidth]{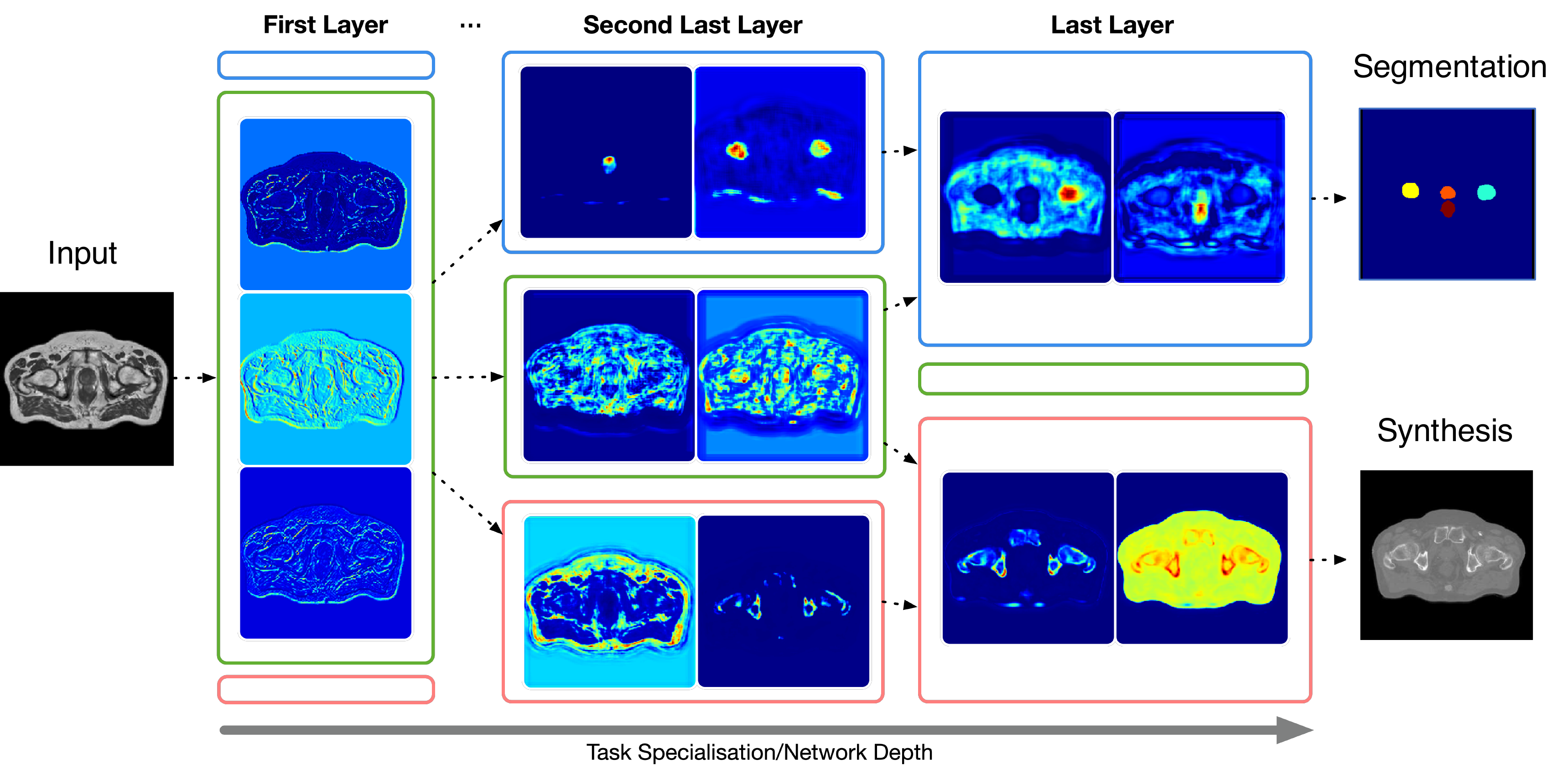}
 	\vspace{-4mm}
	\caption{Activation maps from example kernels in the learned task-specific and shared filter groups, $G^{(l)}_{1}, G^{(l)}_{2}, G^{(l)}_{s}$ (enclosed in blue, green and pink funnels) in the first, the second last and the last convolution layers in the SFG-HighResNet model trained on the medical imaging dataset. The results from convolution kernels with low entropy (i.e. high ``confidence'') of group assignment probabilities $\mathbf{p}^{(l)}$  are shown for the respective layers.  }
    \label{fig:activations}
\end{figure*}

\vspace{-1mm} 
\label{sec:learned_architecture}
Analysis of the grouping probabilities of a network embedded with SFG modules permits visualisation of the network connectivity and thus the learned MTL architecture. To analyse the group allocation of kernels at each layer, we computed the sum of class-wise probabilities per layer. Learned groupings for both SFG-VGG11 network trained on UTKFace and the SFG-HighResNet network trained on prostate scans are presented in Fig.~ \ref{fig:LEARNED_A}. These figures illustrate increasing task specialisation in the kernels with network depth. At the first layer, all kernels are classified as shared (\textbf{p}$=[0, 1, 0]$) as low-order features such as edge or contrast descriptors are generally learned earlier layers. In deeper layers, higher-order representations are learned, which describe various salient features specific to the tasks. This coincides with our network allocating kernels as task specific, as illustrated in Fig.~\ref{fig:activations}, where activations are stratified by allocated class per layer. \textcolor{black}{Density plots of the learned kernel probabilities and trajectory maps displaying training dynamics, along with more examples of feature visualisations, are in Supp.Sec.~C and ~D.} The corresponding results in the case of duplicate tasks (two duplicates of the same task) are also provided in Supp.Sec.~E.

Notably, the learned connectivity of both models shows striking similarities to hard-parameter sharing architectures commonly used in MTL. Generally, there is a set of shared layers, which aim to learn a feature set common to both tasks. Task-specific branches then learn a mapping from this feature space for task-specific predictions. Our models are able to automatically learn this structure whilst allowing asymmetric allocation of task-specific kernels with no priors on the network structure.

\vspace{-1mm} 
\subsection{Effect of \textbf{p} initialisation}
\vspace{-1mm} 
Fig.~\ref{fig:different_grouping} shows the layer-wise proportion of the learned kernel groups on the UTKFace dataset for four different initilization schemes of grouping probabilities $\mathbf{p}$: (i) ``dominantly shared'', with $\mathbf{p}=[0.2, 0.6, 0.2]$, (ii) ``dominantly task-specific'', with $\mathbf{p}=[0.45, 0.1, 0.45]$, (iii) ``random'', where $\mathbf{p}$ is drawn from $\text{Dirichlet}(1,1,1)$, (iv) ``start with MT-constant mask", where an equal number of kernels in each layer are set to probabilities of $\mathbf{p}= [1,0,0],[0,1,0]$ and $[0,0,1]$. In all cases, the same set of hyper-parameters, including the annealing rate of the temperature term in GSM approximation and the coefficient of the entropy regularizer $\mathcal{H}(\mathbf{p})$, were used during training. We observe that the kernel grouping of respective layers in (i), (ii) and (iii) all converge to a very similar configuration observed in Sec.~\ref{sec:learned_architecture}, highlighting the robustness of our method to different initialisations of $\mathbf{p}$. In case (iv), the learning of $p$ were much slower than the remaining cases, due to weaker gradients, and we speculate that a higher entropy regularizer is necessary to facilitate its convergence.

\begin{figure}[b!]
	\center
	\vspace{-7mm}
	\includegraphics[width=0.9\linewidth]{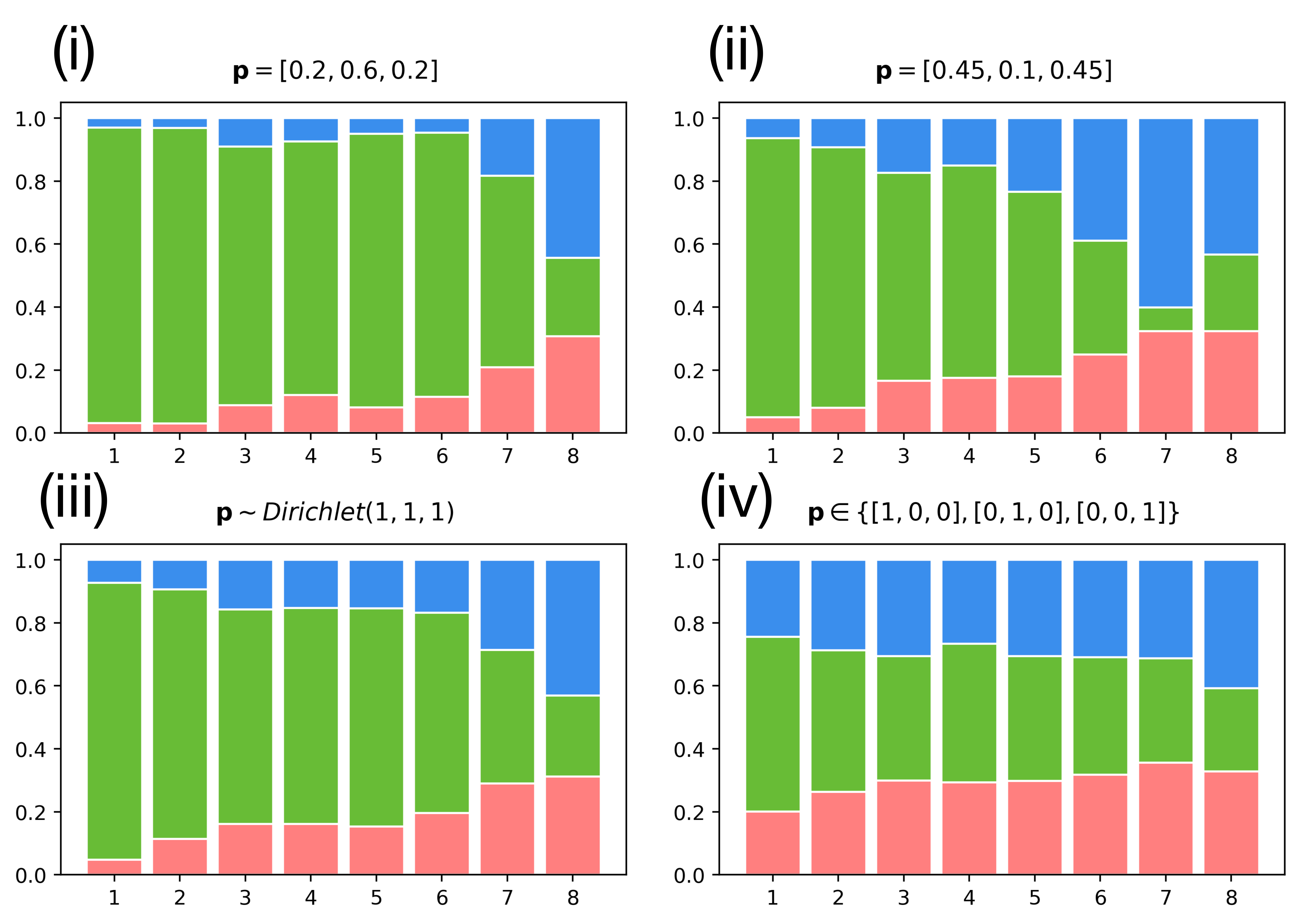}
	\vspace{-3mm}
	\caption{\small Effect of the initial values of grouping probabilities $\mathbf{p}$ on the learned kernel allocation after convergence. }
    \label{fig:performance_on_small_data}
    \vspace{-2mm}
\end{figure}

\section{Discussion}
\vspace{-2mm}
In this paper, we have proposed \emph{stochastic filter groups} (SFGs) to disentangle \emph{task-specific} and \emph{generalist} features. SFGs probabilistically defines the grouping of kernels and thus the connectivity of features in a CNNs. We use variational inference to approximate the distribution over connectivity given training data and sample over possible architectures during training. Our method can be considered as a probabilistic form of multi-task architecture learning \cite{liang2018evolutionary}, as the learned posterior embodies the optimal MTL architecture given the data.\looseness=-1

Our model learns structure in the representations. The learned shared (generalist) features may be exploited either in a transfer learning or continual learning scenario. As seen in \cite{lacost2018}, an effective prior learned from multiple tasks can be a powerful tool for learning new, unrelated tasks. Our model consequently offers the possibility to exploit the learned task-specific and generalist features when faced with situations where a third task is needed, which may suffer from unbalanced or limited training data. This is particularly relevant in the medical field, where training data is expensive to acquire as well as laborious. We will investigate this in further work.\looseness=-1

Lastly, a network composed of SFG modules can be seen as a superset of numerous MTL architectures. Depending on the data and the analysed problem, SFGs can recover many different architectures such as single task networks, traditional hard-parameter sharing, equivalent allocation across tasks, and asymmetrical grouping (Fig.~\ref{fig:different_grouping}). Note, however, that proposed SFG module only learns connectivity between neighbouring layers. Non-parallel ordering of layers, a crucial concept of MTL models \cite{meyerson2018beyond, Ruder2019SluiceNL}, was not investigated. Future work will look to investigate the applicability of SFG modules for learning connections across grouped kernels between non-neighbouring layers.

\subsubsection*{Acknowledgments}
FB and MJC were supported by CRUK Accelerator Grant A21993. RT was supported by Microsoft Scholarship. DA was supported by EU Horizon 2020 Research and Innovation Programme Grant 666992, EPSRC Grant M020533, R014019, and R006032 and the NIHR UCLH BRC. We thank NVIDIA Corporation for hardware donation.

{\small
\bibliographystyle{unsrt}
\bibliographystyle{ieee_fullname}
\bibliography{egbib}
}

\newpage
\appendix

\clearpage

\section{Training and implementation details}
\subsection{Optimisation, regularisation and initialisation}
All networks were trained with ADAM optimiser \cite{ADAM} with an initial learning rate of $10^{-3}$ and $\beta = [0.9, 0.999]$. We used values of $\lambda_1=10^{-6}$ and $\lambda_2=10^{-5}$ for the weight and entropy regularisation factors in Equation $(5)$ in Section $3.2$. All \emph{stochastic filter group} (SFG) modules were initialised with grouping probabilities \textbf{p}=[$0.2$, $0.6$, $0.2$] for every convolution kernel. Positivity of the grouping probabilities $\textbf{p}$ is enforced by passing the output through a \emph{softplus} function $f(x)=\text{ln}(1+e^{x})$ as in \cite{lakshminarayanan2017simple}. The scheduler $\tau = \text{max}(0.10, \exp(-rt))$ recommended in \cite{jang2016categorical} was used to anneal the Gumbel-Softmax temperature $\tau$ where $r$ is the annealing rate and $t$ is the current training iteration. We used $r=10^{-5}$ for our models.

Hyper-parameters for the annealing rate and the entropy regularisation weight were obtained by analysis of the network performance on a secondary randomly split on the UTK dataset ($70/15/15$). They were then applied to all trained models (large and small dataset for UTKFace and medical imaging dataset).

\subsection{UTKFace}
For training the VGG networks (Section 4.1 - UTKFace network), we used the root-mean-squared-error (RMSE) for age regression and the cross entropy loss for gender classification. The labels for age were divided by $100$ prior to training. The input RGB images ($200$x$200$x$3$) were all normalised channel wise to have unit variance and zero mean prior to training and testing. A batch-size of $10$ was used. No augmentation was applied. We monitored performance during training using the validation set ($n=3554$) and trained up to $330$ epochs. We performed $150$ validation iterations every $1000$ iterations, leading to $1500$ predictions per validation iteration. Performance on the validation set was analysed and the iteration where Mean Absolute Error (MAE) was minimised and classification Accuracy was maximised was chosen for the test set.

\subsection{Medical imaging dataset}
We used T2-weighted Magnetic Resonance Imaging (MRI) scans (3T, 2D spin echo, TE/TR: 80/2500ms, voxel size 1.46x1.46x5mm$^{3}$) and Computed Tomography (CT) scans (140 kVp, voxel size 0.98x0.98x1.5 mm$^{3}$). The MR and CT scans were resampled to isotropic resolution (1.46mm$^{3}$). We performed intensity non-uniformity correction on the MR scans \cite{Tustison2010}.

In the HighResNet networks (Section 4.1 - Medical imaging network), we used the RMSE loss for the regression task and the Dice + Cross-Entropy loss \cite{dicex} for the segmentation task. The CT scans were normalised using the transformation $\nicefrac{\text{CT}}{1024} + 1$. The original range of the CT voxel intensity was $[-1024,2500]$ with the background set to $-1024$. The input MRI scans were first normalised using histogram normalisation based on the $1^{st}$ and $99^{th}$ percentile \cite{Nyul2000}. The MRI scans were then normalised to zero mean and unit variance. At test time, input MRI scans were normalised using the histogram normalisation transformation obtained from the training set then normalised to have zero mean and unit variance. 

All scans were of size $288$x$288$x$62$. We sub-sampled random patches from random axial slices of size $128$x$128$. We sampled from all axial slices in the volume ($n=62$). We trained up to $200,000$ iterations using a batch-size of $10$. We applied augmentation to the randomly sampled patches using random scaling factors in the range $[-10\%, 10\%]$ and random rotation angles in the range [$-10^\circ$, $10^\circ$]. The trained patches were zero-padded to increase their size to $136$x$136$. However, the loss during training was only calculated in non-padded regions.

The inference iteration for the test set was determined when the performance metrics on the training set (Mean Absolute Error and Accuracy) first started to converge for at least $10,000$ iterations. In our model where the grouping probabilities were learned, the iteration when convergence in the update of the grouping probabilities was first observed was selected since performance generally increased as the grouping probabilities were updated. 

\subsection{Implementation details}
We used Tensorflow and implemented our models within the NiftyNet framework \cite{Gibson2018}. Models were trained on NVIDIA Titan Xp, P6000 and V100. All networks were trained in the Stochastic Filter Group paradigm. Single-task networks were trained by hard-coding the allocation of kernels to task $1$ and task $2$ i.e. $50$\% of kernels per layer were allocated to task $1$ and $50$\% were allocated to task $2$ with constant probabilities \textbf{p}=[1,0,0] and \textbf{p}=[0,0,1] respectively. The multi-task hard parameter sharing (MT hard-sharing) network was trained by hard-coding the allocation of kernels to the shared group i.e. $100$\% of kernel per layer were allocated to the shared group with constant probability \textbf{p}=[0, 1, 0]. The cross-stitch (CS) \cite{MisraCrossMTL16} networks were implemented in a similar fashion to the single-task networks, with CS modules applied to the output of the task-specific convolutional layers. The other baselines (MT-constant mask and MT-constant \textbf{p}=[\nicefrac{1}{3}, \nicefrac{1}{3}, \nicefrac{1}{3}]) were trained similarly.

We used Batch-Normalisation \cite{ioffe2015batch} to help stabilise training. We observed that the deviation between population statistics and batch statistics can be high, and thus we did not use population statistic at test time. Rather, we normalised using batch-statistics instead, and this consistently lead to better predictive performance. We also used the Gumbel-Softmax approximation \cite{jang2016categorical} at test-time using the temperature value $\tau$ that corresponded to the iteration in $\tau$ annealing schedule.

\section{CNN architectures and details}
\begin{figure*}[h]
    \centering
    \includegraphics[width=0.80\textwidth]{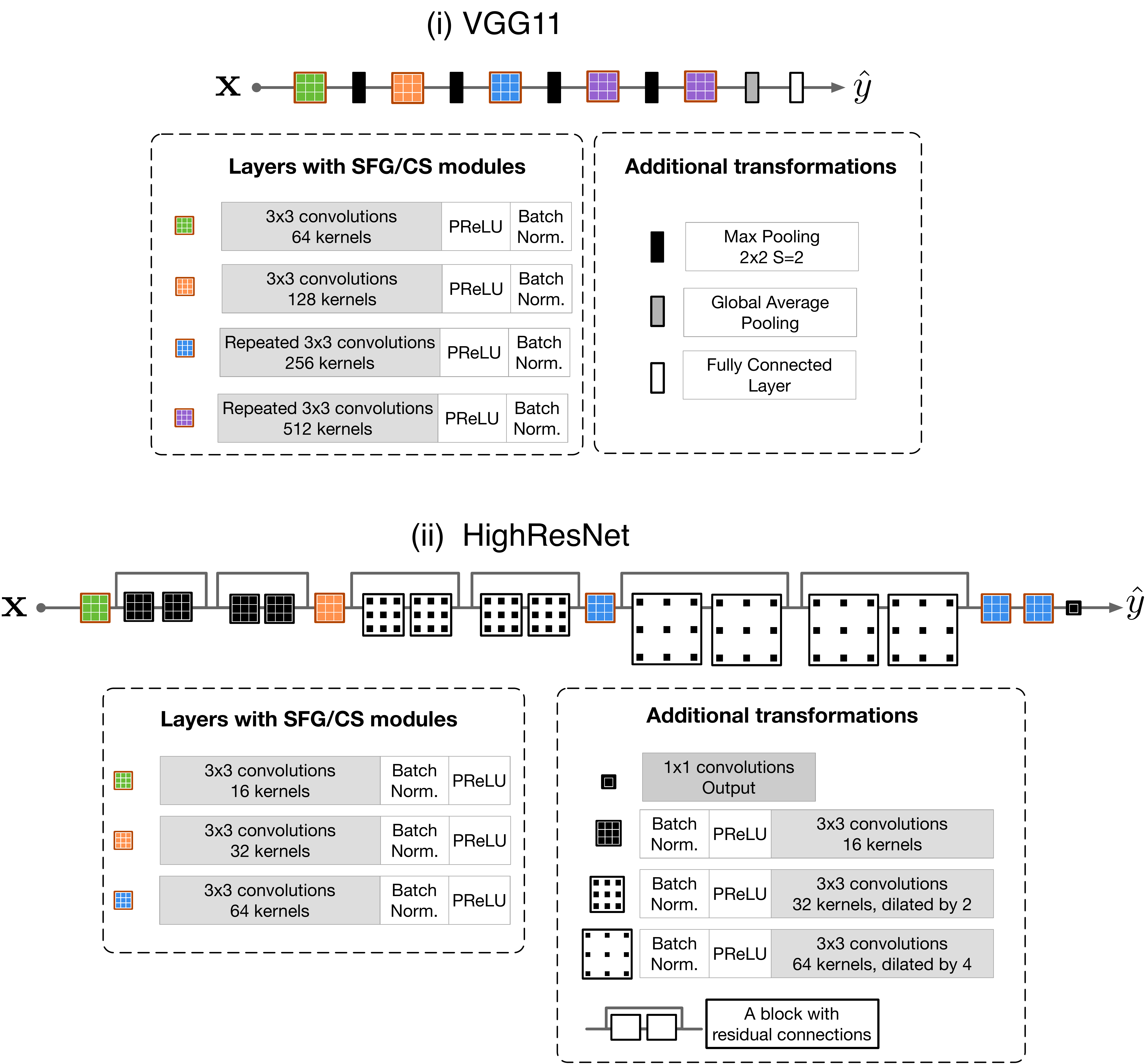}
    \caption{Illustration of the single-task architectures, (i) VGG11 and (ii) HighResNet used for UTKFace and medical imaging dataset, respectively. In each architecture, the coloured components indicate the layers to which SFG or cross-stitch (CS) modules are applied when extended to the multi-task learning scenario, whilst the components in black denote the additional transformations applied to the outputs of respective filter groups or CS operations (see the description of black circles in the schematic provided in Fig.~5 of the main text)}. 
    \label{fig:supp_baselines}
\end{figure*}

We include schematics and details of the single-task VGG11 \cite{vgg} and HighResNet \cite{wenqi} networks in Fig.~\ref{fig:supp_baselines}. In this work, we constructed multi-task architectures by augmenting these networks with the proposed SFG modules. We used the PReLU activation function \cite{prelu} in all networks. For the residual blocks used in the HighResNet networks in Fig.~\ref{fig:supp_baselines} (ii), we applied PReLU and batch-norm as pre-activation \cite{preactivation} to the convolutional layers. The SFG module was used to cluster the kernels in every coloured layer in Fig.~\ref{fig:supp_baselines}, and distinct sets of additional transformations (pooling operations for VGG and high-res blocks for HighResNet) were applied to the outputs of the respective filter groups $G_{1}, G_{2}, G_{s}$. For a fair comparison, the CS units \cite{MisraCrossMTL16} were added to the same set of layers.

For clarification, the SFG layer number $n$ (e.g. SFG layer 2) corresponds to the $n^{th}$ layer with an SFG module. In the case of SFG-VGG11, each convolutional layer uses SFGs. The SFG layer number thus corresponds with layer number in the network. In the case of SFG-HighResNet, not every convolutional layer uses SFGs such as those within residual blocks. Consequently, SFG layer 1 corresponds to layer 1, SFG layer 2 is layer 6, SFG layer 3 is layer 11, SFG layer 4 is layer 16 and SFG layer 5 is layer 17.

\section{Learned grouping probability plots}
\begin{figure*}[ht!]
    \vspace{-2mm}
    \centering
    \includegraphics[width=0.9\textwidth]{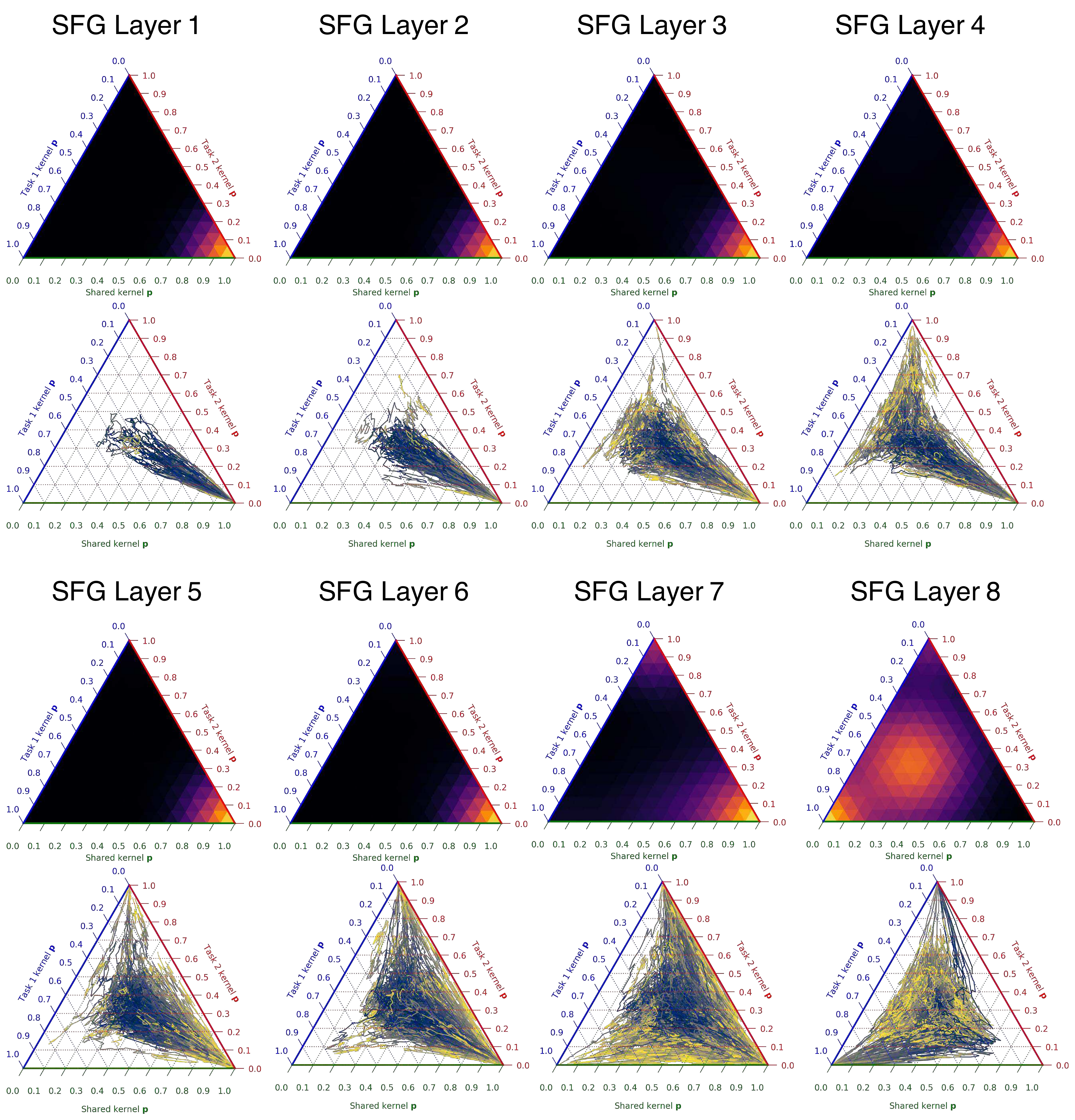}
    \caption{Density plots and trajectory plots of the learned grouping probabilities for the SFG-VGG11 architecture. The density plots represents the final learned probabilities per layer for each kernel. The trajectory plots represent how the grouping probabilities are learned during training and thus how the connectivity is determined. Histograms of the grouping probabilities were smoothed with a Gaussian kernel with $\sigma=1$. The densities are mapped to and visualised in the $2$-simplex using \texttt{python-ternary} \cite{ternary}.}
    \label{fig:fig_learned_A}
\end{figure*}

\begin{figure*}[ht]
    \centering
    \vspace{-5mm}
    \includegraphics[width=1.0\textwidth]{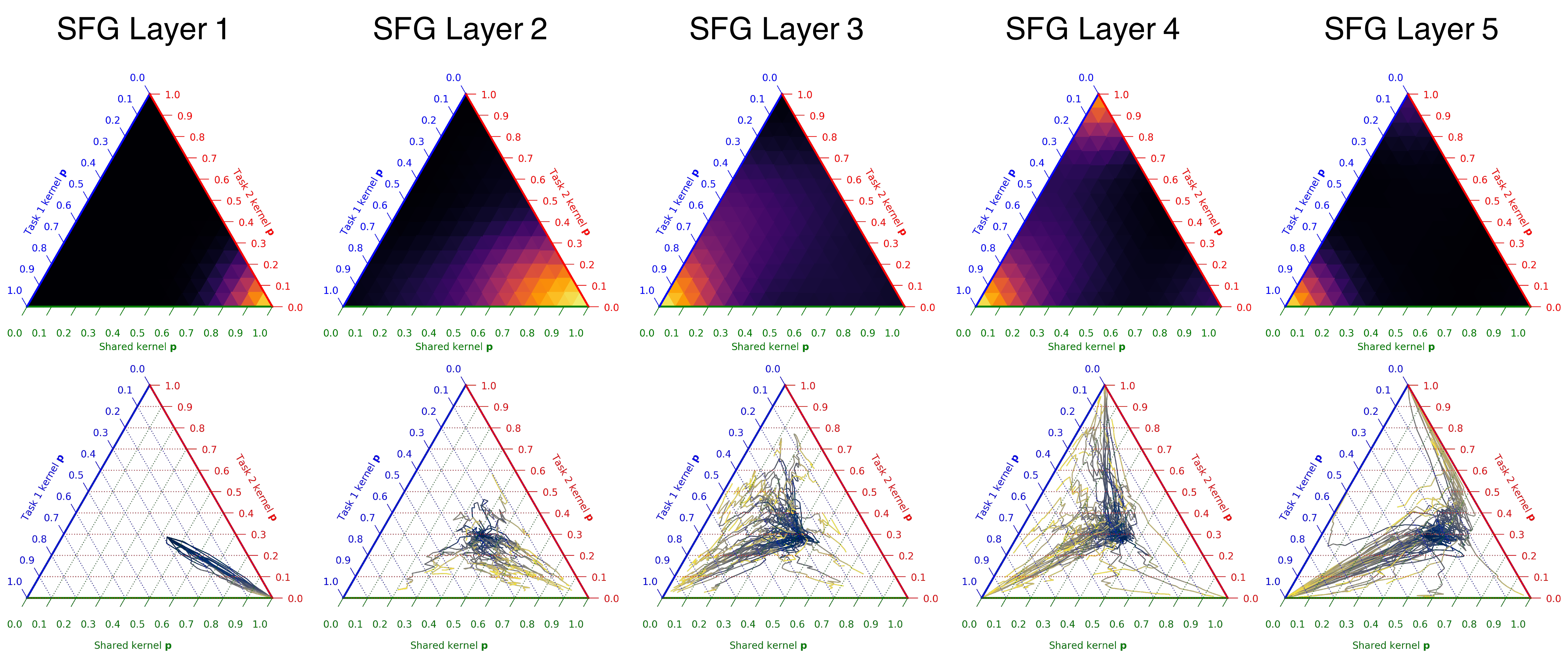}
    \caption{Density plots and trajectory plots of the learned grouping probabilities for the SFG-HighResNet architecture. The density plots represents the final learned probabilities per layer for each kernel. The trajectory plots represent how the grouping probabilities are learned during training and thus how the connectivity is determined.}
    \label{fig:fig_learned_B}
    \vspace{-5mm}
\end{figure*}

In this section, we illustrate density plots of the learned grouping probabilities $\mathbf{p}$ for each trained network (Fig.~\ref{fig:fig_learned_A} and Fig.~\ref{fig:fig_learned_B}). We also plot the training trajectories of grouping probabilities $\mathbf{p}$ of all kernels in each layer. These are colour coded by iteration number---blue for low and yellow for high iteration number. This shows that some grouping probabilities are quickly learned in comparison to others. 

Fig.~\ref{fig:fig_learned_A} and Fig.~\ref{fig:fig_learned_B} show that most kernels are in the shared group at earlier layers of the network where mostly low-order generic features are learned (as illustrated in Fig.~ \ref{fig:act}, SFG layer $1$). They converge quickly to the shared vertex of the $2$-simplex as evidenced by the colour of the trajectory plots. As the network depth increases, task-specialisation in the kernels increases (see Fig.~ \ref{fig:act}, SFG layer $\ge4$). This is illustrated by high density clusters at task-specific vertices and by the trajectory plots.

\section{Extra visualisation of activations}
\begin{figure*}[h]
    \centering
    \vspace{-4mm}
    \includegraphics[width=0.8\textwidth]{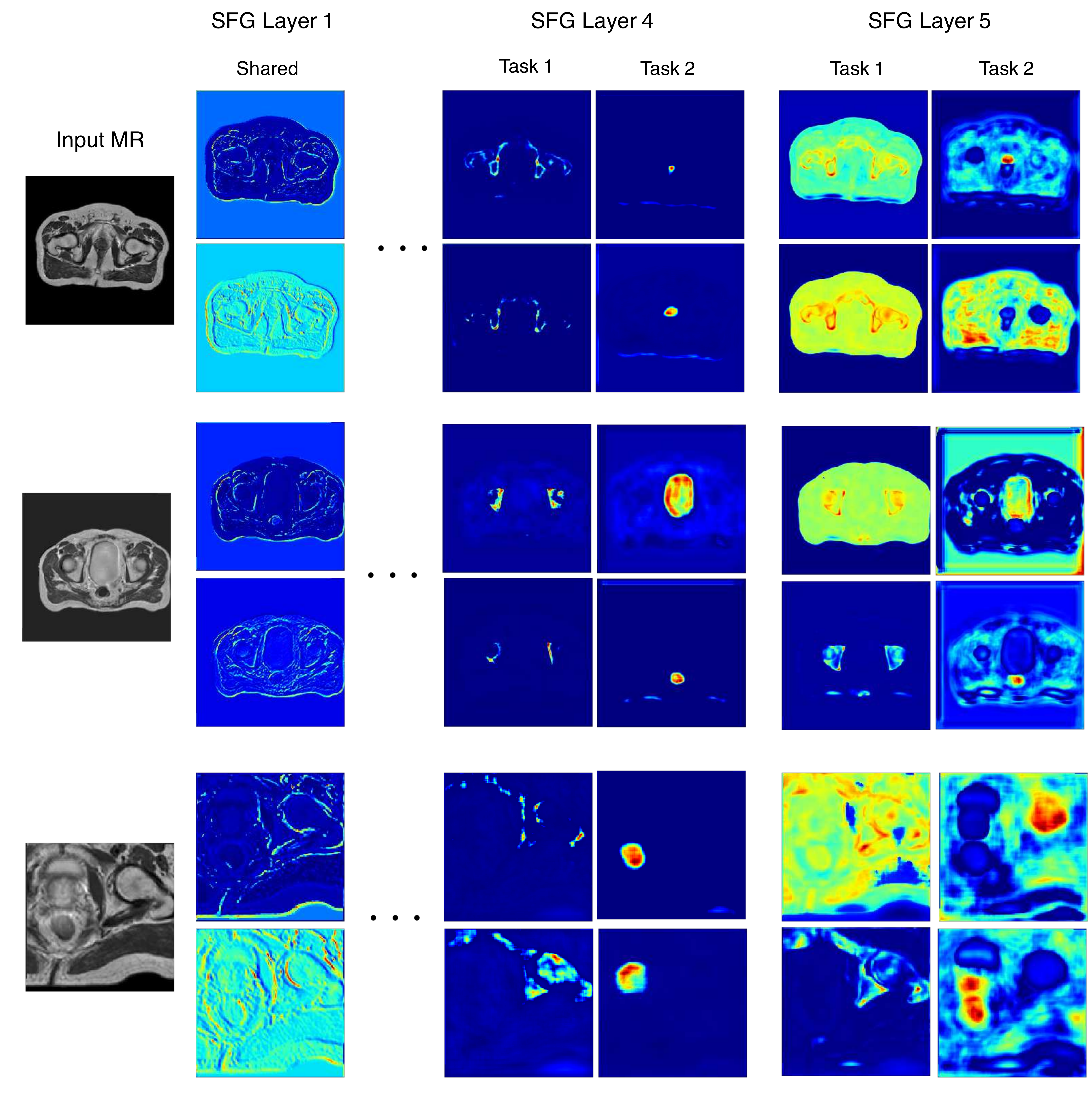}
    \vspace{-2mm}
    \caption{Example activations for kernels with low entropy of $\mathbf{p}$ (i.e. group assignment with high confidence) for three input MR slices in the SFG-HighResNet multi-task network. Columns ``Shared'', ``Task 1'' \& ``Task 2'' display the results from the shared, CT-synthesis and organ-segmentation specific filter groups in respective layers. We illustrate activations stratified by group in layer $1$ (SFG layer 1), layer $16$ (SFG layer 4) and layer $17$ (SFG layer 5). }
    \label{fig:act}
    \vspace{-1mm}
\end{figure*}

\begin{figure*}[h]
    \centering
    \includegraphics[width=0.85\textwidth]{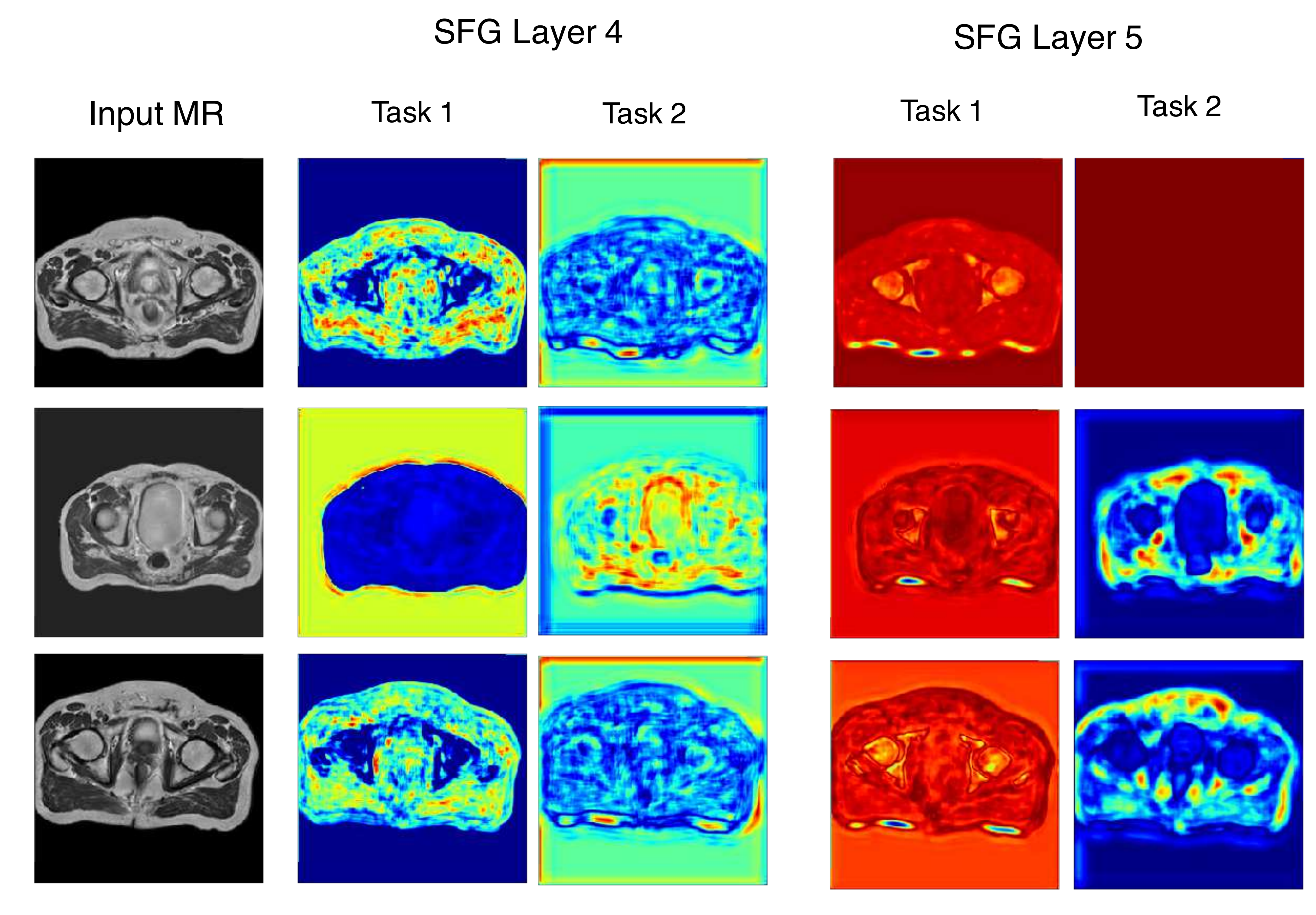}
     \caption{Example activations for kernels with high entropy (i.e. group assignment with low confidence) for three input MR slices in the SFG-HighResNet multi-task network.  Columns ``Shared'', ``Task 1'' \& ``Task 2'' display the results from the shared, CT-synthesis and organ-segmentation specific filter groups in respective layers. We illustrate activations stratified by group in layer $16$ (SFG layer 4) and layer $17$ (SFG layer 5). }
    \label{fig:act_unc}
\end{figure*}

Here we visualise the activation maps of additional specialist and generalist kernels on the medical imaging dataset. To classify each kernel according to the group (task 1, task 2 or shared), we selected the group with the respective maximum assignment probability. The corresponding activation maps for various input images in the medical imaging dataset can be viewed in Fig.~\ref{fig:act} and Fig.~\ref{fig:act_unc}. 

We first analysed the activation maps generated by kernels with low entropy of $\mathbf{p}$ (i.e. highly confident group assignment). At the first layer, all kernels are classified as shared, and the examples in Fig.~\ref{fig:act} support that these kernels tend to account for low-order features such as edge and contrast of the images. On the other hand, at deeper layers, higher-order representations are learned, which describe various salient features specific to the tasks such as organs for segmentation, and bones for CT-synthesis. Note that the bones are generally the most difficult region to synthesise CT intensities from an input MR scan \cite{bragman2018multi}.

Secondly, we looked at activation maps from kernels with high entropy of $\mathbf{p}$ (i.e. highly uncertain group assignment) in Fig.~\ref{fig:act_unc}. In contrast to Fig.~\ref{fig:act}, the learned features do not appear to capture any meaningful structures for both synthesis and segmentation tasks. Of particular note is the dead kernel in the top row of the figure; displaying that a high uncertainty in group allocation correlates with non-informative features. 

\section{Learned filter groups on duplicate tasks}
\begin{figure*}[h]
    \centering
    \includegraphics[width=1.0\textwidth]{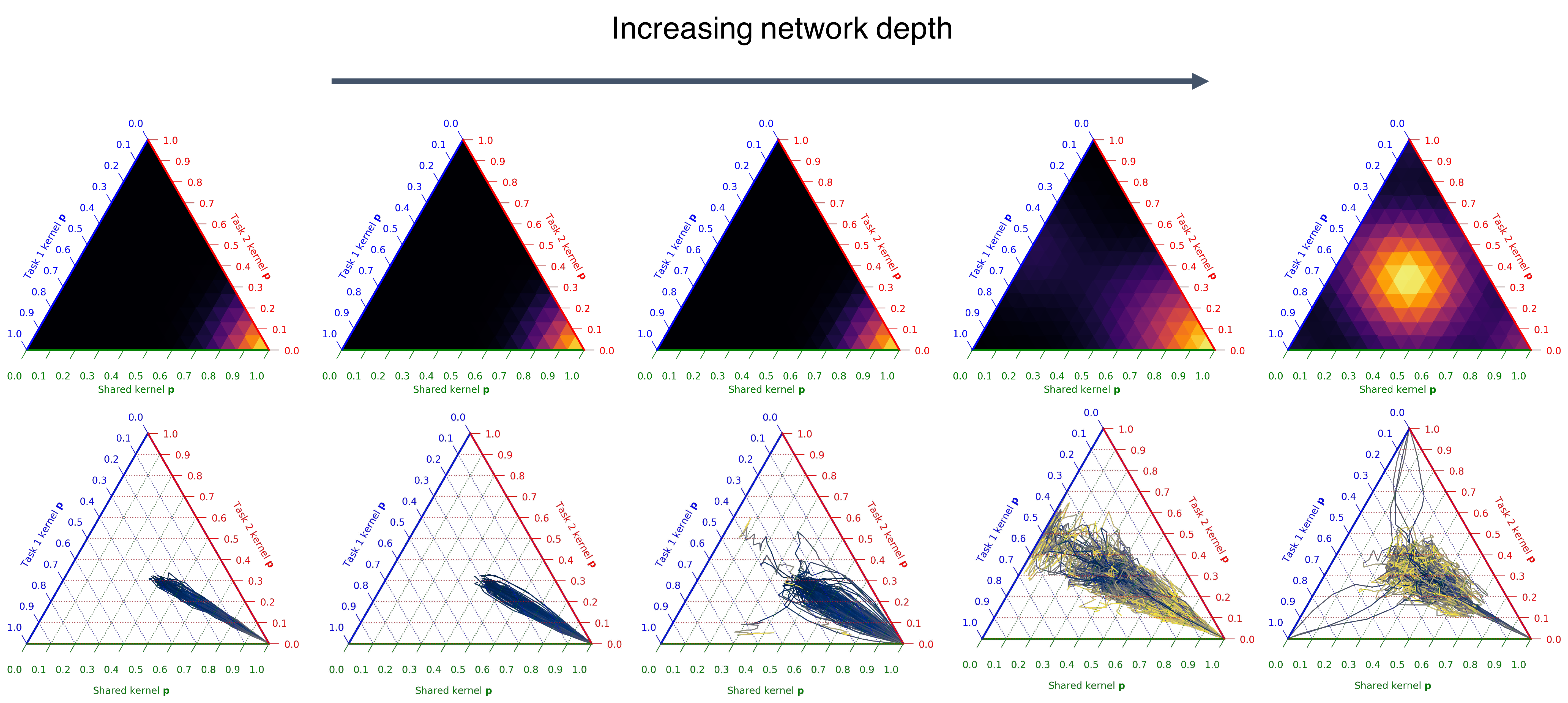}
    \caption{Top: density plots for the learned grouping probabilities at each SFG layer in a model where we trained on duplicate tasks i.e. task 1 is CT synthesis and task 2 is also CT synthesis. Bottom: trajectories of the grouping probabilities during training.}
    \label{fig:duplicate}
\end{figure*}

We analysed the dynamics of a network with SFG modules when trained with two duplicates of the same CT regression task (instead of two distinct tasks). Fig.~\ref{fig:duplicate} visualises the learned grouping and trajectories of the grouping probabilities during training. In the first $3$ SFG layers (layers $1$, $6$ and $11$ of the network), all the kernels are grouped as shared. In the penultimate SFG layer (layer $16$), either kernels are grouped as shared or with probability \textbf{p}=[\nicefrac{1}{2}, 0, \nicefrac{1}{2}], signifying that the kernels can belong to either task. The final SFG layer (layer $17$) shows that most kernels have probabilities \textbf{p}=[\nicefrac{1}{3}, \nicefrac{1}{3}, \nicefrac{1}{3}]. Kernels thus have equal probability of being task-specific or shared. This is expected as we are training on duplicate tasks and therefore the kernels are equally likely to be useful across all groups.

\end{document}